\documentclass{article}

\usepackage[preprint, nonatbib]{templates/neurips/neurips_2026}
\makeatletter\renewcommand{\@notice}{}\makeatother

\usepackage[numbers]{natbib}

\usepackage[utf8]{inputenc} %
\usepackage[T1]{fontenc}    %
\usepackage{url}            %
\usepackage{booktabs}       %
\usepackage{amsfonts}       %
\usepackage{nicefrac}       %
\usepackage{microtype}      %
\usepackage{cite}
\usepackage{amsmath,amssymb,amsfonts}
\usepackage{graphicx}
\usepackage{textcomp}
\usepackage{nicefrac}
\usepackage{pifont}
\usepackage{xcolor}
\usepackage{mathabx}
\usepackage{fontawesome5}

\usepackage{soul}
\usepackage{caption}
\usepackage[table]{xcolor}
\usepackage{anyfontsize}
\usepackage[normalem]{ulem}
\usepackage{wrapfig}
\usepackage{multirow}
\usepackage{tabularx}
\usepackage[export]{adjustbox}
\usepackage{stackengine}
\usepackage{float}
\usepackage[edges]{forest}
\usepackage{array}
\usepackage{stfloats}
\usepackage{xspace}
\usepackage[outline]{contour}

\usepackage{makecell}
\usepackage{subcaption}
\usepackage{pgfplots}
\usepackage{tabularray}
\usepackage[para]{footmisc}
\usepackage{fancyvrb}
\usepackage{placeins}
\usepackage{tikz-3dplot}
\usepackage{pgf-pie}
\usetikzlibrary{calc, shapes, patterns, shadings, positioning, arrows.meta,3d, intersections}
\usepackage{styles}
\usepgfplotslibrary{fillbetween,colormaps}
\pgfplotsset{compat=1.17}
\usepackage[colorlinks,bookmarks,pagebackref,breaklinks]{hyperref}
\usepackage{bookmark}
\usepackage{paralist}
\usepackage{balance}

\usepackage[skip=2pt]{caption}
\usetikzlibrary{tikzmark,patterns,shapes.misc,shapes.geometric,positioning,calc}

\usepackage[capitalize]{cleveref}
\crefname{section}{Sec.}{Secs.}
\Crefname{section}{Section}{Sections}
\Crefname{table}{Table}{Tables}
\crefname{table}{Tab.}{Tabs.}

\title{UniverSat: Resolution- and Modality-Agnostic Transformers for Earth Observation}

\author{
  \textbf{Yohann Perron\textsuperscript{$\ast$,1,5}} \quad
  \textbf{Guillaume Astruc\textsuperscript{$\ast$,1,2,4}} \quad
  \textbf{Nicolas Gonthier\textsuperscript{2,3}} \\
  \textbf{Clément Mallet\textsuperscript{2}} \quad
  \textbf{Loic Landrieu\textsuperscript{1,2}} \\[0.3cm]
  \textsuperscript{1}LIGM, Ecole Nationale des Ponts et Chaussées, IP Paris, Univ Gustave Eiffel, CNRS\\
  \textsuperscript{2}LASTIG, Univ Gustave Eiffel, IGN, ENSG,
  \quad
  \textsuperscript{3}IGN \quad
  \textsuperscript{4}CNES \quad
  \textsuperscript{5}EFEO 
}

\makeatletter
\renewcommand{\section}{%
  \@startsection{section}{1}{\z@}%
  {-2.0ex plus -0.5ex minus -0.2ex}%
  {0.3ex plus 0.1ex}%
  {\normalfont\Large\bfseries}%
}

\renewcommand{\subsection}{%
  \@startsection{subsection}{2}{\z@}%
  {-1.5ex plus -0.4ex minus -0.2ex}%
  {0.3ex plus 0.1ex}%
  {\normalfont\large\bfseries}%
}
\makeatother

\begin{document}
\setlength{\leftmargini}{2em}
\tdplotsetmaincoords{60}{30}

\setcounter{footnote}{0}

\maketitle

\begin{figure}[H]
\begin{minipage}{0.35\linewidth}
   \caption{\textbf{One Model, Many Sensors.}
A single \textsc{UniverSat} is trained jointly on $13$ sensors from $7$ datasets, with wide variations in spatial resolution, channel count, and revisit frequency.
The resulting model generalizes to unseen sensors within this gamut \emph{without input resampling}.}
    \label{fig:relaycount}
\end{minipage}
\hfill
\begin{minipage}{0.60\linewidth}
    \centering
    \resizebox{\linewidth}{!}{
    \hspace{-5mm}
    \definecolor{colVHR}{RGB}{46,178,164}%
\definecolor{colOpt}{RGB}{254,170,94}%
\definecolor{colRadar}{RGB}{110,150,230}%
\definecolor{colHyper}{RGB}{214,106,205}%
\definecolor{colDEM}{RGB}{205,205,205}%
\begin{tabular}{c}
\begin{tikzpicture}
\begin{axis}[
  width=\linewidth, height=5.5cm,
  axis lines=left,                 %
  axis line style={->, very thick},
  xmin=0.7, xmax=5.35,
  ymin=0,   ymax=1,
  zmin=0.7, zmax=5.3,
  enlargelimits=false,
  clip=false,
  x dir=normal,
  y dir=normal,
  z dir=normal,
  x post scale=1,
  y post scale=0.6,      %
  z post scale=1,
  view={15}{15},
  xmajorgrids=true,
  zmajorgrids=true,
  ymajorgrids=true,
  tick label style={font=\small},   %
  label style={font=\small},        %
  xtick pos=lower,  ytick pos=left,  ztick pos=left,
  xtick align=outside, ytick align=outside, ztick align=outside,
  tick style={thick},
  xtick={1,2,3,4,5},
  xticklabels={0.1\,m, 1\,m, 10\,m, 30\,m, 250\,m},
  ytick={0,0.3,0.6,0.9},
  yticklabels={none, yearly, weekly, daily},
  yticklabel style={rotate=-25, anchor=west, inner sep=1pt, yshift=-5pt},
  ztick={1,2,3,4,5},
  zticklabels={1, 3, 7, 10, 400},
  xlabel={\textbf{Spatial resolution}},
  ylabel={\textbf{Revisit time}},
  zlabel={\textbf{Number of channels}},
  xlabel style={at={(ticklabel cs:.5)}, yshift=+10pt, rotate=-2},
  ylabel style={at={(ticklabel cs:.5)}, yshift=+10pt, xshift=-40pt,  rotate=5},
  zlabel style={at={(ticklabel cs:.5)}, yshift=-5pt, rotate=0},
]

\EllipticExtrusionXZ{1.18}{2.25}{0.3}{0.3}{0.0}{0.0}{colVHR}
\node[anchor=center, text=colVHR] at ([canvas is xz plane at y=0] 1.5,1.4) {\footnotesize \shortstack{VHR \\ UHR}};

\EllipticExtrusionXZ{3}{4}{0.3}{0.3}{0.0}{0.4}{colOpt}
\node[anchor=east, text=colOpt] at ([canvas is xz plane at y=0] 2.8,4.2) {\footnotesize Sentinel-2};

 \EllipticExtrusionXZ{3}{1.25}{0.3}{0.3}{0.0}{0.0}{colDEM}
 \node[anchor=west, text=black!60] at ([canvas is xz plane at y=0] 3.2,1.25) {\footnotesize DSM, DTM};

\EllipticExtrusionXZ{5}{3.5}{0.3}{0.3}{0.0}{0.6}{colOpt}
 \node[anchor=north, text=colOpt] at ([canvas is xz plane at y=0] 5.5,3.3) {\footnotesize \shortstack{MODIS}};

\EllipticExtrusionXZ{3.5}{2}{0.3}{0.3}{0.0}{0.4}{colRadar}
 \node[anchor=east, text=colRadar] at ([canvas is xz plane at y=0] 5.3,2) {\footnotesize \shortstack{Sent-1, \\ ALOS-2}};

\EllipticExtrusionXZ{4}{2.75}{0.3}{0.3}{0.0}{0.4}{colOpt}
 \node[anchor=east, text=colOpt] at ([canvas is xz plane at y=0] 3.8,3) {\footnotesize \shortstack{Landsat \\ 7,8,9}};

\EllipticExtrusionXZ{4}{5}{0.3}{0.3}{0.0}{0.0}{colHyper}
 \node[anchor=west, text=colHyper] at ([canvas is xz plane at y=0] 4.2,5) {\footnotesize \shortstack{Gaofen-5 \\ EO-1}};

\EllipticExtrusionXZ{2}{5}{0.3}{0.3}{0.0}{0.2}{colHyper}
 \node[anchor=east, text=colHyper] at ([canvas is xz plane at y=0] 1.8,5) {\footnotesize NIS};

\EllipticExtrusionXZ{2}{2.2}{0.3}{0.3}{0.0}{0.2}{colOpt}
\node[anchor=south, text=colOpt] at ([canvas is xz plane at y=0] 2.0,2.8) {\footnotesize  \shortstack{SPOT-6/7}};

  \end{axis}
\end{tikzpicture}
\\[-3mm]
\begin{tabular}{c c c}
    \small \textcolor{colVHR}{\bf aerial} 
    &
    \small \textcolor{colOpt}{\bf optical satellite} 
    &
    \small \textcolor{colRadar}{\bf radar} 
    \\
    \multicolumn{3}{c}{
    \begin{tabular}{c c}
    \small \textcolor{colHyper}{\bf hyperspectral} 
    &
    \small \textcolor{colDEM}{\bf elevation}
    \end{tabular}
    }
\end{tabular}
\end{tabular}
    }
\end{minipage}
\label{fig:teaser}
\end{figure}

\begin{abstract}
Vision Transformers (ViT) dominate computer vision. However, their reliance on rigid patch projectors hinders transfer to Earth Observation (EO), where input modalities, scales, and resolutions vary widely. We introduce \textit{UniverSat}, a ViT-style backbone built around a \emph{Universal Patch Encoder} that maps patches from arbitrary spatial, spectral, and temporal resolutions, and from both optical and non-optical sensors, into a shared embedding space with a shared set of weights. This enables training a single model on heterogeneous multimodal corpora via self-supervision, yielding robust, sensor-agnostic spatial features. We validate this approach with strong results across classification and segmentation on standard EO benchmarks from GeoBench, PANGEABench, and SpectralEarth. Our code and models are available at \href{https://github.com/gastruc/UniverSat}{\nolinkurl{github.com/gastruc/UniverSat}}.
\end{abstract}

\blfootnote{\small $^{\ast}$Equal contribution.}

\section{Introduction}
\vspace{-0mm}

Vision Transformers (ViTs) are the de facto backbone for visual representation learning, owing to their versatility and  compatibility with self-supervised learning (SSL) objectives: masking~\citep{he2022masked}, contrastive learning~\citep{he2020momentum}, predictive methods~\citep{assran2023self}, and clustering-based approaches~\citep{caron2021emerging,oquab2023dinov2}.
This paradigm has been successfully applied to Earth Observation (EO)~\citep{tseng2025galileo,astruc2024anysat,cong2022satmae,jakubik2025terramind,brown2025alphaearth}.
However, EO is more than ``just vision'': it spans heterogeneous \emph{modalities} (optical, radar, LiDAR, elevation), \emph{scales} (from hectares to tens of km$^2$), and \emph{resolutions}: spatial (cm--km), temporal (single acquisitions to dense time series), and spectral (1--400 bands). This diversity is a strength only with an architecture able to exploit it.

Most EO foundation models adopt the ViT architecture with minimal adaptation, implicitly fixing band counts, temporal sampling, and patch size.
When operating on a narrow set of widely available modalities (\eg, Sentinel or aerial RGB imagery), this design is appropriate and largely sufficient.
However, EO practitioners often combine sensors with widely varying pixel sizes, spectral and temporal densities, and spatial extents. Under these conditions, a rigid architecture becomes limiting and often requires aggressive resampling, band selection/aggregation, or modality-specific processing.
While recent work improves robustness across scales~\citep{reed2023scale,astruc2024omnisat}, datasets~\citep{astruc2024anysat,waldmann2025panopticon}, and spectral resolution~\citep{prexl2024senpa,sumbul2025smarties,xiong2024dofa,houdre2025ramen}, most approaches remain constrained to specific modality sets and rely on modality-specific projectors. We argue that this rigidity, inherited from fixed ViT-style patch projectors, limits the generality of EO foundation models.

We introduce \textsc{UniverSat}, a ViT-like model tailored to EO by replacing fixed patch projectors with a \emph{Universal Patch Encoder} (UPE).
The UPE uses linear-complexity axial cross-attention to embed patches of arbitrary spatial, spectral, and temporal dimensions into a shared latent space.
Integrated into a transformer operating over spatialized tokens, this design provides four key advantages:
\vspace{-1.5mm}
\begin{compactitem}
\item \textbf{Modality-Agnostic.} A single set of weights processes many modalities combinations and arbitrary resolutions without input resampling or channel filtering.
\item \textbf{Resolution-flexible.} The spatial resolution of the output feature map is specified at inference time and decoupled from the input patch size.
\item \textbf{Granular.} A skip connection preserves fine spatial details beyond patch-level embeddings.
\item \textbf{SSL-compatible.} The architecture retains the core ViT structure, enabling training with standard self-supervised learning  objectives.
\end{compactitem}

We train \textsc{UniverSat} with a self-supervised objective that extends recent latent masked modeling~\citep{yi2023masked,herzog2025olmoearth} to multimodal and multitemporal EO data.
A single model is trained jointly on $7$ datasets spanning $13$ sensors across four modalities (optical, hyperspectral, radar, elevation), covering $0.1$--$300$\,m Ground Sampling Distance (GSD), $1$--$150$ timestamps, and $1$--$396$ spectral channels.
Despite its flexibility and ability to incorporate unseen sensor configurations, \textsc{UniverSat} remains highly competitive on standard benchmarks, as demonstrated through linear and convolutional probing on GeoBench~\citep{lacoste2023geo}, PANGEABench~\citep{marsocci2024pangaea}, and challenging hyperspectral benchmarks~\citep{braham2025spectralearth}.
Our main contributions are as follows:\vspace{-2mm}
\begin{compactitem}
\item A unified ViT-like architecture for EO that processes heterogeneous sensors without modality-specific projectors or preprocessing;
\item A multimodal self-supervised training framework tailored to heterogeneous EO data;
\item Competitive performance across a broad range of datasets and tasks, from high-resolution RGB imagery to challenging radar time series and hyperspectral benchmarks;
\item Demonstrated generalization to unseen sensors (EnMAP, PlanetScope) and modality combinations.
\end{compactitem}

\section{Related work}
\vspace{-0mm}
\begin{table*}[t]
    \centering
    \caption{\textbf{Flexible Multimodal EO Foundation Models.}
For each model, we list the training modalities, whether unseen spatial/temporal/spectral configurations are handled \emph{without input resampling}, and the feature-map granularity.
\textsc{UniverSat} supports the broadest modality mix, handles unseen configurations, and offers flexible output resolution---all with a single set of weights.}
    \resizebox{1\linewidth}{!}{
    \definecolor{colVHR}{RGB}{46,178,164}%
\definecolor{colOpt}{RGB}{254,170,94}%
\definecolor{colRadar}{RGB}{110,150,230}%
\definecolor{colHyper}{RGB}{214,106,205}%
\definecolor{colDEM}{RGB}{205,205,205}%
\begin{tabular}{cc}
\begin{tabular}{l m{5mm}m{5mm}m{5mm}m{5mm}m{5mm} m{10mm}m{10mm}m{10mm} m{20mm}}
\toprule
 & \multicolumn{5}{c}{training modalities}
 & \multicolumn{3}{c}{supports unseen resolution}
 & \multirow{2}{*}[-2mm]{\makecell{output\\resolution}}
 \\ \cmidrule(lr){2-6}\cmidrule(lr){7-9}
 & 
\cellcolor{colVHR!60}{\faCamera}& \cellcolor{colOpt!60} 
 \faVideo & \cellcolor{colRadar!60} 
 \faRadar{1} & \cellcolor{colDEM!60} 
 \faElevation{1} & \cellcolor{colHyper!60} 
 \faHyper{1}
 & spat. & temp. & spec. & 
 \\\midrule
\rowcolor{gray!10}
TerraMind~\citep{jakubik2025terramind} & \applycolorA{1} & & \applycolorC{1} &\applycolorE{1} & & \faCheck & & & \faPatch{1}\\ %
\rowcolor{gray!0}OmniSat~\citep{astruc2024omnisat} & \applycolorA{2} & \applycolorB{2} & \applycolorD{1} & \applycolorE{1} & & & \faCheck & &\faFullImage{1} \\
\rowcolor{gray!10}EarthView~\citep{velazquez2025earthview} & \applycolorA{3}  &   & \applycolorC{1}  &\applycolorF{1} & & & \faCheck & &\faPatch{1}\\
\rowcolor{gray!0}Panopticon~\citep{waldmann2025panopticon} & \bf \applycolorA{5} & & \applycolorC{1} & & \applycolorF{1} & & & \faCheck & \faPatch{1} \\
\rowcolor{gray!10}Galileo~\citep{tseng2025galileo} &  & \applycolorB{1}  &\applycolorD{1} & \applycolorE{1}& &  \faCheck  & \faCheck & &\faPatch{1}\\
\rowcolor{gray!0}AnySat~\citep{astruc2024anysat} &  \applycolorA{2} &  \bf \applycolorB{4}  &\bf \applycolorD{2} &  \applycolorE{1} & & \faCheck & \faCheck & & \faPixel{1}\\
\rowcolor{gray!10}DOFA~\citep{xiong2024dofa} & \applycolorA{3} &  & \applycolorC{1}  & &\applycolorF{1} & \faCheck & & \faCheck  &\faPatch{1}\\
\rowcolor{gray!0}FoMo~\citep{bountos2025fomo}& \applycolorA{4} & \applycolorB{2} &\applycolorD{1} & \applycolorE{1} &  & & \faCheck & \faCheck & \faPatch{1} \\ 
\rowcolor{gray!10}Ramen~\citep{houdre2025ramen} & \applycolorA{1} & \applycolorB{1} & \applycolorC{1} & \applycolorD{1} & &  \faCheck & \faCheck & \faCheck & \faAnyRes{1} \\ 
\greyrule 
\rowcolor{gray!0}\bf UniverSat (ours) &  \applycolorA{4}  & \bf \applycolorB{4} & \bf \applycolorD{2} & \bf \applycolorE{2} & \bf \applycolorG{3} & \faCheck& \faCheck  & \faCheck & \faAnyRes{1}
\\ \bottomrule
\end{tabular}
&
\begin{tabular}{r@{ : }l}
\faCamera & optical snapshot \\
\faVideo & optical time series \\
\faRadar{1} & radar \\
\faElevation{1} & elevation \\
\faHyper{1} & hyperspectral \\[2mm]
\multicolumn{2}{l}{output resolutions:} \\
\faFullImage{1} & image \\
\faPatch{1} & patch \\
\faPixel{1} & pixel \\
\faAnyRes{1} & selectable 
\end{tabular}
\end{tabular}

\if10
\\
\resizebox{\linewidth}{!}{
\begin{tabular}{c}
\begin{tabular}{r@{ : }l r@{ : }l r@{ : }l r@{ : }l r@{ : }l r@{ : }l r@{ : }l}
\faCamera/\faVideo & optical snapshot/ time series & \faRadar{1} & radar & \faElevation{1} & elevation & \faHyper{1} & hyperspectral
\end{tabular}\\
\begin{tabular}{l r@{ : }l r@{ : }l r@{ : }l r@{ : }l  }
output resolutions: & \faFullImage{1} & image & \faPatch{1} & patch &  \faPixel{1} & pixel &  \faAnyRes{1} & selectable 
\end{tabular}
\end{tabular}
}
\fi

    }
    \label{tab:models}
\end{table*}

We provide an overview of the state-of-the-art in flexible, multimodal foundation models for Earth Observation (EO); a synthesis is presented in \cref{tab:models}, and an extended list  in the appendix.

\paragraph{Self-supervised Learning for EO.}
Early EO SSL was predominantly contrastive~\citep{ayush2021geography,manas2021seasonal,mall2023change,tseng2022croco, feng2025tessera}.
Subsequent work explored generative masking in the spirit of MAE ~\citep{cong2022satmae,reed2023scale,han2024bridging,Mendieta2023}, and predictive objectives~\citep{tseng2025galileo,astruc2024anysat}.
Several methods incorporate EO priors---spectral~\citep{cong2022satmae,sumbul2025smarties}, temporal~\citep{manas2021seasonal,dumeur2024self}, and spatial~\citep{reed2023scale,ayush2021geography}--either in the loss or in the encoder.
Nevertheless, many pipelines remain monomodal/monotemporal, which limits their generality.
We also note that some SSL settings include semantic products as inputs, effectively moving toward semi-supervision~\citep{tseng2025galileo,brown2025alphaearth,tseng2023lightweight,danish2025terrafm,zhu2025skysense}.

\paragraph{Multimodal SSL for EO.}
Cross-modal pretraining leverages geo-registration: different sensors observe the same scene in a common reference frame.
Masked cross-modal reconstruction \citep{astruc2024omnisat,labatie2025maestro,wang2022self} and cross-sensor contrast \citep{tseng2022croco, fuller2023croma, astruc2026unigeoclip} have both been effective.
Teacher–student hybrids add stability at scale~\citep{brown2025alphaearth,zhu2025skysense,zhang2025skysense}.
In practice, most studies focus on Sentinel-1/2 {time-series} and single-date inputs~\citep{prexl2024senpa,fuller2023croma}.
Some models also apply to elevation
\citep{astruc2024anysat,jakubik2025terramind,astruc2024omnisat,bountos2025fomo,fayad2025dunia} and hyperspectral data~\citep{waldmann2025panopticon,xiong2024dofa}.

\paragraph{Flexible Models.}
Many EO backbones already support variable-length time series~\citep{astruc2024anysat,brown2025alphaearth,labatie2025maestro,zhang2025skysense}.
Spectral variability is typically handled with token-per-band strategies~\citep{prexl2024senpa,bountos2025fomo}, band grouping~\citep{tseng2025galileo,cong2022satmae,irvin2023usat}, or continuous wavelength encodings and dynamic weights~\citep{waldmann2025panopticon,xiong2024dofa,sumbul2025smarties,houdre2025ramen}.
Spatial generalization at inference is addressed through feature pyramids~\citep{xiong2024dofa,weber2025pyvit}, scale-aware positional encodings~\citep{reed2023scale}, or FlexiViT-style encoders~\citep{tseng2025galileo,houdre2025ramen,herzog2025olmoearth,beyer2023flexivit}; nevertheless, many pipelines still resample inputs, inflating data volume by orders of magnitude.
A persistent bottleneck is the \emph{fixed patch projector}: it enforces rigid input formats (mono\-temporal/fixed patch size) or modality-specific encoders~\citep{astruc2024anysat,tseng2025galileo}, and typically requires retraining for new configurations.
\emph{Atomizer}~\citep{de2025atomizer} and \emph{Ramen}~\citep{houdre2025ramen} are conceptually close to our work with per-pixel, per-band ``atoms'', but spatially resample its inputs to accommodate varying resolutions.
To the best of our knowledge, \textsc{UniverSat} is the first EO foundation model to jointly support flexibility along the spatial, spectral, and temporal axes without input resampling, see \cref{tab:models}.

\section{Method}
\label{sec:Method}
\vspace{-0mm}

We introduce a method to process and merge heterogeneous Earth observations into dense feature maps, independently of their respective resolutions.
At its core is a \emph{Universal Patch Encoder} (UPE) that projects patches of arbitrary shape into a shared embedding space (\cref{sec:upe}).
Built on top of UPE, \textsc{UniverSat} aggregates multimodal observations into dense spatial representations (\cref{sec:archi}), and is trained in a self-supervised manner on a heterogeneous multimodal corpus (\cref{sec:training}).
Each component addresses one consequence of EO heterogeneity: the UPE maps patches of any shape to a fixed width, with axial attention keeping this tractable; cross-modal contrast aligns the UPE outputs of co-registered sensors; ACA fuses modalities at the patch level; a spatial Transformer models interactions between patches; masked prediction on fixed random targets trains the whole encoder with targets of uniform size across sensors; and a sub-patch skip connection restores fine spatial detail for dense prediction.

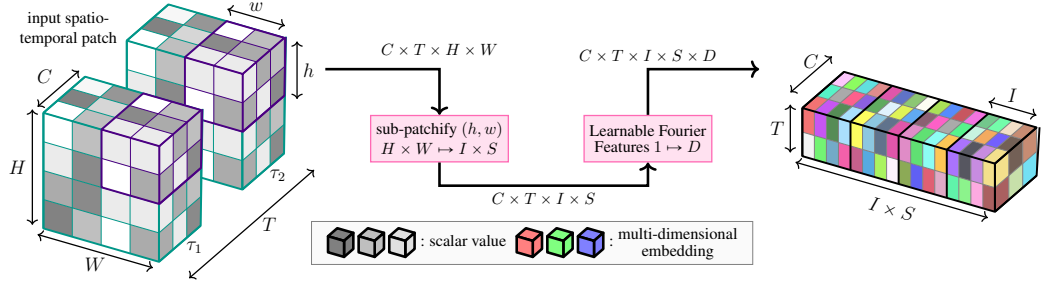
\begin{figure}[t]
    \def\patchsize{2.4}%
\def\subsize{1.2}%
\def\pixelsize{0.6}%
\def\xlegend{8}%
\def\ylegend{-2.25}%
\def\xreshape{6.0}%
\def\xFourier{10.5}%
\def\xout{16}%
\def\yreshape{0}%
\def\yout{0}%
\def\yFourier{0}%
\def\channelC{1.5}%
\def\xshift{0}%
\def\yshift{+6}%
\def\zshift{0}%
\def\xshiftcnn{-6}%
\def\zshifttwo{-8}%
\def\gap{0.5}%
\definecolor{PATCHCOLOR}{RGB}{0, 150, 136}%
\definecolor{SUBCOLOR}{RGB}{75,0,130}%
\definecolor{PIXELCOLOR}{RGB}{255,102,0}%
\definecolor{ATOMCOLOR}{RGB}{186,0,148}%
\definecolor{CONVCOLOR}{RGB}{75,0,130}%
\definecolor{LINCOLOR}{RGB}{255, 105, 180}%
\resizebox{1\linewidth}{!}{
\begin{tikzpicture}

\node[draw=none] (input) at (0,0) 
{
\begin{tikzpicture}[tdplot_main_coords, line join=round,scale=0.6]

\draw [thick, black, <->] (-\patchsize,-\channelC-\gap,-\patchsize) -- node [draw=none, below] {\large $W$} (+\patchsize,-\channelC-\gap,-\patchsize);

\draw [thick, black, <->] (+\xshift,+\channelC+ \yshift,+\patchsize+\zshift+\gap) -- node [draw=none, above] {\large $w$} (+\patchsize+\xshift,+\channelC+ \yshift,\patchsize+\zshift+\gap);

\draw [thick, black, <->] (-\patchsize,-\channelC-\gap,+\patchsize+\gap) -- node [draw=none, above left] {\large $C$} (-\patchsize,\channelC-\gap,+\patchsize+\gap);

\draw [thick, black, <->] (-\patchsize-\gap,-\channelC,-\patchsize) -- node [draw=none, left] {\large $H$} (-\patchsize-\gap,-\channelC,\patchsize);

\draw [thick, black, <->] (+\patchsize+\gap+\xshift,+\channelC+\yshift,0+\zshift) -- node [draw=none, right] {\large $h$} (+\patchsize+\gap+\xshift,+\channelC+\yshift,\patchsize+\zshift);

\node [draw=none, text=black] at (+\patchsize+1,-\channelC/2,-\patchsize-\gap+1) {\large $\tau_1$};
\node [draw=none, text=black] at 
(+\patchsize+1+\xshift,-\channelC/2+\yshift,-\patchsize-\gap+\zshift+1) {\large $\tau_2$};

\draw [thick, black, <->] (+\patchsize+1.5,-\channelC-\gap,-\patchsize) -- node [draw=none, right=1mm] {\large $T$} (+\patchsize+\xshift+1.5,+\channelC-\gap+\yshift,-\patchsize+\zshift);

\node[draw=none] at (0+\xshift,+\yshift,+\zshift)
{
\begin{tikzpicture}[tdplot_main_coords, line join=round,scale=0.6]
\cubenotikzbigedges{4}{3}{4}{\patchsize*2}{\channelC*2}{\patchsize*2}{1}{PATCHCOLOR}{0}
\end{tikzpicture}
};

\node[draw=none] at (\subsize+\xshift,+\yshift,\subsize+\zshift)
{
\begin{tikzpicture}[tdplot_main_coords, line join=round,scale=0.6]
\cubenotikzbigedges{2}{3}{2}{\subsize*2}{\channelC*2}{\subsize*2}{1}{SUBCOLOR}{0}
\end{tikzpicture}
};

\node[draw=none] at (0,0,0)
{
\begin{tikzpicture}[tdplot_main_coords, line join=round,scale=0.6]
\cubenotikzbigedges{4}{3}{4}{\patchsize*2}{\channelC*2}{\patchsize*2}{1}{PATCHCOLOR}{0}
\end{tikzpicture}
};

\node[draw=none] at (\subsize,0,\subsize)
{
\begin{tikzpicture}[tdplot_main_coords, line join=round,scale=0.6]
\cubenotikzbigedges{2}{3}{2}{\subsize*2}{\channelC*2}{\subsize*2}{1}{SUBCOLOR}{0}
\end{tikzpicture}
};

\end{tikzpicture}
};

 \node[draw=LINCOLOR, fill=LINCOLOR!20, thick, minimum height= 1cm] (reshape) at (\xreshape, \yreshape)
 {\shortstack{sub-patchify $(h,w)$\\
 $H\times W \mapsto I \times S$
 }};

\node[draw=LINCOLOR, fill=LINCOLOR!20, thick, minimum height= 1cm] (Fourier) at (\xFourier, \yFourier)
{\shortstack{Learnable Fourier\\ Features  $1 \mapsto D$}
};

\node [draw=none] (output) at (\xout, \yout) 
{
\begin{tikzpicture}[tdplot_main_coords, line join=round,scale=0.6]

\draw [thick, black, <->] (0,-0.5,0) -- node [draw=none, below] {\large \rotatebox{-15}{$I \times S$}} (8,-0.5,0);

\draw [thick, black, <->] (-0.5,0,2.5) -- node [draw=none, above] {\large $C$} (-0.5,3,2.5);

\draw [thick, black, <->] (-0.5,0,+0) -- node [draw=none, left] {\large $T$} (-0.5,0,2);

\draw [thick, black, <->] (6,3,2.5) -- node [draw=none, above] {\large $I$} (8,3,2.5);

\cubenotikz{16}{3}{2}{8}{3}{2}{5}{black}{1}
\end{tikzpicture}
};

\draw [ultra thick, ->] (input.east) ++ (0,1.5) -| node [pos=0.5] (elbow) {} (reshape);

\node[draw=none, above=0mm of elbow] {$C \times T \times H \times W$};

\draw [ultra thick, ->] (reshape.south)  -- ++ (0,-0.5) -| node [pos=0.25, below] {$C \times T \times I \times S$} (Fourier.south);

 \draw [ultra thick, ->] (Fourier.north)
 -- node[pos=1, above] {$C \times T \times I \times S \times D$} (Fourier.north |- elbow) -- (output.west |- elbow);

\node[draw=black!50, fill=black!2] at (\xlegend, \ylegend)
{
\begin{tabular}{c@{\,}c@{\,}c@{\;:\;}l@{\;\;}c@{\,}c@{\,}c@{\;:\;}l}
    \begin{tikzpicture}[baseline=-.5ex, scale=0.2, tdplot_main_coords]
  \simplecubetopfill{0}{0}{0}{1}{1}{1}{black!50}
  \end{tikzpicture}%
  &
  \begin{tikzpicture}[baseline=-.5ex, scale=0.2, tdplot_main_coords]
  \simplecubetopfill{0}{0}{0}{1}{1}{1}{black!25}
  \end{tikzpicture}%
  &
  \begin{tikzpicture}[baseline=-.5ex, scale=0.2, tdplot_main_coords]
   \simplecubetopfill{0}{0}{0}{1}{1}{1}{black!12}
  \end{tikzpicture}%
  &
  scalar value
 &
    \begin{tikzpicture}[baseline=-.5ex, scale=0.2, tdplot_main_coords]
  \simplecubetopfill{0}{0}{0}{1}{1}{1}{red!50}
  \end{tikzpicture}%
  &
  \begin{tikzpicture}[baseline=-.5ex, scale=0.2, tdplot_main_coords]
  \simplecubetopfill{0}{0}{0}{1}{1}{1}{green!50}
  \end{tikzpicture}%
  &
  \begin{tikzpicture}[baseline=-.5ex, scale=0.2, tdplot_main_coords]
   \simplecubetopfill{0}{0}{0}{1}{1}{1}{blue!50}
  \end{tikzpicture}%
   & 
   \raisebox{-2mm}{\shortstack{multi-dimensional \\ embedding}}
  \end{tabular}
};
\node [draw=none, above left= -13mm and -27mm of input]  {\shortstack{input spatio-\\ temporal patch}};

\end{tikzpicture}
}
    \caption{\textbf{Patch Formatting.}
    An input patch of size $C \times T \times H \times W$ is converted to a tensor of size  $C \times T \times I \times S$  with $S$ sub-patch of $I=h\,w$ pixels. We then lift all scalar values to dimension $D$ with learned Fourier features.
 }
    \label{fig:atoms}
\end{figure}

\subsection{Universal Patch Encoder}
\label{sec:upe}

The Universal Patch Encoder (UPE) replaces the fixed patch projector of a standard ViT.
Its goal is to produce a fixed-width embedding from a patch acquired by any sensor or modality, regardless of its spatial, spectral, or temporal resolution.
To achieve this, UPE represents patches as atomic tokens and progressively collapses their dimensions using axial cross-attention \citep{ho2019axial}.

\para{Atomic formatting.}
We consider a patch
$
x \in \bR^{C \times T \times H \times W},
$
with $C$ channels, $T$ timestamps, and spatial extent $H\times W$. Channels may encode heterogeneous measurements across modalities, including spectral bands (optical), polarization responses (radar), or elevation products (\eg, DSM/DTM).
We factorize the spatial grid into $S = HW/(hw)$ subpatches of size $I = hw$ pixels, and reshape $x$ as $C \times T \times I \times S$.
Each scalar value $x_{c,t,i,s}$ is then lifted to dimension $D$ using Learnable Fourier Features (LFF, see Appendix for details), yielding
\begin{align}
e = \mathrm{LFF}(x),
\quad
\text{dim: } C \times T \times I \times S \times D~.
\end{align}
We refer to these $D$-dimensional vectors as \emph{atomic tokens}~\citep{de2025atomizer}.
Note that each atom is already a scalar (one value for one pixel, channel, and timestamp): LFF \emph{expands} it rather than compressing a vector.
LFF is cheap enough to apply to every atom while being nonlinear, which lets the network form value-selective responses (\eg, band-pass or threshold-like filters on reflectance, backscatter, or elevation); a linear scalar embedding performs worse (\cref{tab:ablation}).

\begin{figure*}[t]
    \centering
    \def\xsymb{-3}
\def\xinput{-0}
\def\xpool{3.5}
\def\xlinear{5.5}
\def\xkey{5}
\def\xvalue{7.5}
\def\xquerry{4}
\def\xca{8}
\def\xres{8.5}
\def\xFF{10.0}
\def\xplus{13.0}
\def\xout{12}

\def\yinput{0}
\def\ymid{1}
\def\yquerry{2}

\def\linwidth{1.5}
\def\CAheight{1}
\def\pluswidth{0.5}
\def\keysep{0.5}
\def\shift{0.5}
\def\acaheight{1.5}
\def\acawidth{1.5}
\def\keygap{0.5}

\def\expansion{1.7}

\def\bigstep{0}

\definecolor{LINCOLOR}{RGB}{255, 105, 180}
\definecolor{ACACOLOR}{RGB}{0, 150, 136}
\definecolor{TENSORCOLOR}{RGB}{255, 105, 180}
\definecolor{OPERATIONCOLOR}{RGB}{153, 102, 255}

\resizebox{\linewidth}{!}{
\begin{tikzpicture}
[
linear/.style={draw=LINCOLOR, fill=LINCOLOR!20, rectangle, rounded corners, align=center, very thick,inner sep=1mm, minimum width=\linwidth cm},
tensor/.style={draw=TENSORCOLOR, fill=TENSORCOLOR!20, rectangle,  align=center, very thick,inner sep=1mm, minimum width=\linwidth cm},
operation/.style={draw=OPERATIONCOLOR, fill=OPERATIONCOLOR!20, ellipse,  align=center, very thick,inner sep=1mm, minimum width=\linwidth cm},
aca/.style={draw=ACACOLOR, fill=ACACOLOR!20, rectangle, rounded corners, minimum height=\acaheight cm, minimum width=\acawidth cm, align=right, very thick,inner sep=1mm,text depth = \acaheight cm},
]

\node [draw=none] at (\xinput,\yinput+1.35) {};

\node [aca] (acasymb) at (\xsymb,\ymid) {\footnotesize \shortstack{$\ACA_{X}^\alpha$}};

\node [draw=none, inner sep=0mm] (cubeC) at (\xsymb,\ymid) {
\begin{tikzpicture}[tdplot_main_coords, scale=0.5]
\cubenotikz{1}{1}{1}{1}{1}{1}{0.5}{black}{3}
\begin{pgfinterruptboundingbox}
\draw [draw=red, very thick, ->] (\xmax+\shift,0,0) -- (\xmax+\shift,\ymax,0);
\end{pgfinterruptboundingbox}
\end{tikzpicture}
};

\node [draw=none, right=0mm of acasymb] {\Huge $\mathbf{=}$};

\node [draw=none] (input) at (0, \ymid) 
{
\begin{tikzpicture}[tdplot_main_coords, line join=round,scale=0.6,remember picture, scale=0.75]

\draw [thick, black, <->] (0,-0.5,0) -- node [draw=none, below] {\large \rotatebox{-15}{$Y_1$}} (4,-0.5,0);

\draw [thick, black, <->] (-0.5,0,2.5) -- node [draw=none, above left] {\large \textcolor{red}{$X$}} (-0.5,3,2.5);

\draw [thick, black, <->] (-0.5,0,+0) -- node [draw=none, left] {\large $Y_2$} (-0.5,0,2);

\cubenotikz{4}{3}{2}{4}{3}{2}{5}{black}{1}

\draw [ultra thick, draw=red!100!black, -] (4,3,2) -- (3,3,2) -- (3,0,2) -- (4,0,2) -- (4,3,2) -- (4,3,1) -- (4,0,1) -- (3,0,1) -- (3,0,2) (4,0,1) -- (4,0,2);

\end{tikzpicture}
}; 

\node [draw=none, below=-1mm of input] {\scriptsize $\displaystyle{\textcolor{red}{\mathbf{X}} \times Y \times D}$};

\node [operation] (pool) at (\xpool,\yquerry) { \texttt{pool}$_{X}$};

\node[linear] (linearkey) at (\xlinear,\yinput) {\texttt{Linear}};

\node[linear] (linearquerry) at (\xlinear,\yquerry) {\texttt{Linear}};

\node [draw=none, above=0mm of linearkey] {\footnotesize $D \mapsto 2 \alpha D$};
\node [draw=none, below=0mm of linearquerry] {\footnotesize $D \mapsto \alpha D$};

\node[linear, minimum height=\CAheight cm] (CA) at (\xca,\yquerry) {\texttt {\shortstack{cross-\\attention}}};

\node[linear] (FF) at (\xFF,\yquerry) {\texttt{\shortstack{Feed\\Forward}}};

\node [draw=none, scale=0.75] (output) at (\xout, \yinput+0.75) 
{
\begin{tikzpicture}[tdplot_main_coords, line join=round,scale=0.6]

\draw [thick, black, <->] (0,-0.5,0) -- node [draw=none, below] {\large \rotatebox{-15}{$Y_1$}} (4,-0.5,0);

\draw [thick, black, <->] (-0.5,0,+0) -- node [draw=none, left] {\large $Y_2$} (-0.5,0,2);

\cubenotikz{4}{1}{2}{4}{2}{2}{5}{black}{1}

\draw [ultra thick, draw=red!100!black, -] (4,2,2) -- ++ (-1,0,0) -- ++ (0,-2,0) -- ++ (1,0,0) -- ++ (0,2,0) -- ++ (0,0,-1) -- ++ (0,-2,0) -- ++ (-1,0,0) -- ++ (0,0,+1) (4,0,2) -- ++ (0,0,-1);

\end{tikzpicture}
};   

\node [draw=none, below=-1mm of output] {\scriptsize $\displaystyle{Y \times \alpha D}$};

\node [draw=none, below left= 0mm and 0mm of CA.south] {K,V};
\node [draw=none, above left= 0mm and 0.5mm of CA.west] {Q};

\node [draw=none] (output) at (\xout,\ymid) {};

\draw [draw=none] (input.east) ++ (-0.37,0.45) coordinate (source);

\draw [black, very thick, ->] (source) |- (pool) coordinate[pos=0.85] (elbow) -- (linearquerry) -- (CA);

\draw [black, very thick, ->] (elbow) |- (linearkey) -| (CA.south);

\draw [draw=none] (output.north) ++ (1,-0.15) coordinate (target);

\draw [black, very thick, ->] (CA.east) -- (FF) -| (target);

\end{tikzpicture}
}
    \vspace{-3mm}
   \caption{{\bf Axial Cross-Attention.} Given an input of dimension $X\!\times\!Y$ tokens of dimension $D$ ($Y$ can be several axes), the $\ACA_X^\alpha$ module collapses target dimension $X$ and expands the feature dimension by $\alpha$.
   We first pool along the target dimension $X$ and generate queries $Q$ for the remaining indices. We then
   compute keys $K$ and values $V$ of dimension $\alpha D$ for all tokens and perform cross-attention along dimension $X$, broadcasting along all other indices. Finally, a feed-forward network $\FF$ yields an array of size $Y$ tokens of dimension $\alpha D$.}
    \label{fig:aca}
\end{figure*}
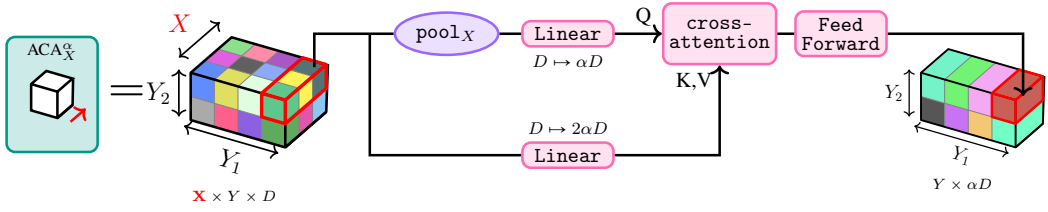

\para{Axial Cross-Attention (ACA).}
Directly projecting $e$ with an MLP is impractical, as $C, T, I,$ and $S$ vary widely across sensors.
Applying full self-attention over all atomic tokens would also be prohibitively expensive, since each axis can reach the hundreds.
Instead, we aggregate tokens one axis at a time using axial cross-attention~\citep{ho2019axial} (see \cref{fig:aca}).

Given a tensor
$t \in \mathbb{R}^{X \times Y \times D}$,
where $X$ is the axis to collapse and $Y$ groups the remaining dimensions, the module $\mathrm{ACA}_X^\alpha$ produces a representation of size $Y \times \alpha D$ by attending over $X$ only.
Concretely, for each index in $Y$, a query is formed from pooled features, while keys and values are computed from all elements along $X$. Attention is then applied along axis $X$, followed by a feed-forward network. Restricting attention to a single axis ensures linear complexity in the number of atomic tokens. Axis-specific metadata is injected before attention with dedicated positional encodings representing the nature of the modality and the characteristics of the atom: wavelength, polarization ratio, time in year, etc.
Full details are provided in the appendix.

\para{Sequential dimension collapse.}
Starting from a patch of shape
$C\!\times\!T\!\times\!I\!\times\!S\!\times\!D$,
we progressively aggregate information by collapsing one axis at a time using ACA.
We follow a fixed order: pixel within each sub-patch ($I$), channel ($C$), time ($T$), and finally sub-patch within the patch ($S$), gradually reducing the tensor while increasing the feature dimension.
Collapsing $I$ first keeps the cost manageable for high-resolution inputs, while keeping $S$ until the end provides the sub-patch embeddings used for dense prediction.
Collapsing channels before time lets temporal attention operate on spectrally contextualized features, which helps with atmospheric and illumination variations; the reverse order degrades performance on long time series (\cref{tab:ablation}):
\begin{align}
\UPE(x) &= (f,\; f^{\text{sub}})~\;\;\text{with}\;\;
f^{\text{sub}} =
\mathrm{ACA}_{T}^2\big(
\mathrm{ACA}_{C}^2\big(
\mathrm{ACA}_{I}^1(e)
\big)\big)~,
\quad
f = \mathrm{ACA}_{S}^2\big(f^{\text{sub}}\big)~.
\end{align}
The UPE produces:
(i) a global patch embedding $f \in \mathbb{R}^{8D}$, and
(ii) sub-patch embeddings $f^{\text{sub}} \in \mathbb{R}^{S \times 4D}$,
which are later used in a high-resolution skip connection.
Both are then linearly projected to the width of the spatial Transformer. The expansion factors above correspond to \textsc{UniverSat}-Tiny; \textsc{UniverSat}-B also expands the pixel axis ($\ACA_I^2$), see the appendix.

\begin{figure*}[t]
    \centering
    \resizebox{\linewidth}{!}{
    \def\xinput{-4.5}%
\def\xacaI{-2.25}%
\def\xpostI{-0.5}%
\def\xacaC{1.25}%
\def\xpostC{3}%
\def\xacaT{4.75}%
\def\xpostT{6.5}%
\def\xacaS{8.25}%
\def\xpostS{10}%
\def\xcubeS{11.75}%
\def\ymain{0}%
\def\ydim{-1.25}%
\def\posencsize{0.8}%
\def\acaheight{1.5}%
\def\acawidth{1.5}%
\def\bigstep{0}%
\def\shift{0.5}%
\def\bigstep{0}%
\def\upshift{0.5}%
\definecolor{ACACOLOR}{RGB}{0, 150, 136}%
\definecolor{SKIPCOLOR}{RGB}{75, 0, 130}%
\begin{tikzpicture}
[
aca/.style={draw=ACACOLOR, fill=ACACOLOR!20, rectangle, rounded corners, minimum height=\acaheight cm, minimum width=\acawidth cm, align=right, very thick,inner sep=1mm,text depth = \acaheight cm},
]
\node [draw=none, inner sep=0mm] (inputcube) at (\xinput,\ymain) {
\begin{tikzpicture}[tdplot_main_coords, scale=0.5]
\def\bigstep{2}
\cubenotikz{4}{4}{4}{2}{2}{2}{0.5}{black}{2}

\node[draw=none] at (-\shift,0,\zmax/2) {\footnotesize T};
\node[draw=none] at (\xmax+\shift,\ymax/2,0) {\footnotesize C};

\node[draw=none] at (\xmax/2,0,-\shift) {\footnotesize \rotatebox{-7}{\!\!\!$I \times S$}};

\end{tikzpicture}};

\node  [draw=none] at (\xinput,\ydim) {
\scriptsize $C \times T \times I \times S \times D$};

\node [aca] (acaI) at (\xacaI,\ymain) {\footnotesize \shortstack{$\ACA_I^1$}};

\node  [draw=none, above=0mm of acaI]  {
\scriptsize $C \times T \times \textcolor{red}{I} \times S \times D$};

\node [draw=none, inner sep=0mm] (cubeI) at (\xacaI,\ymain) {
\begin{tikzpicture}[tdplot_main_coords, scale=0.5]
\cubenotikz{1}{1}{1}{1}{1}{1}{0.5}{black}{3}
\begin{pgfinterruptboundingbox}
\draw [draw=red, very thick, ->] (0,0,-\shift) -- (\xmax/2,0,-\shift);
\draw [draw=red, very thick, ->] (\xmax/2,0,-\shift) -- (\xmax,0,-\shift);
\end{pgfinterruptboundingbox}
\end{tikzpicture}
};

\node [draw=none, inner sep=0mm] (postI) at (\xpostI,\ymain) {
\begin{tikzpicture}[tdplot_main_coords, scale=0.5]
\cubenotikz{2}{4}{4}{1}{2}{2}{0.5}{black}{1}
\end{tikzpicture}
};

\node [draw=none] at (\xpostI,\ydim) {\scriptsize  $C \times T \times S \times D$};

\node [aca] (acaC) at (\xacaC,\ymain) {\footnotesize \shortstack{$\ACA_C^2$}};

\node  [draw=none, above=0mm of acaC]  {
\scriptsize $\textcolor{red}{C}\times T \times S \times D$};

\node [draw=none, inner sep=0mm] (cubeC) at (\xacaC,\ymain) {
\begin{tikzpicture}[tdplot_main_coords, scale=0.5]
\cubenotikz{1}{1}{1}{1}{1}{1}{0.5}{black}{3}
\begin{pgfinterruptboundingbox}
\draw [draw=red, very thick, ->] (\xmax+\shift,0,0) -- (\xmax+\shift,\ymax,0);

\end{pgfinterruptboundingbox}
\end{tikzpicture}
};

\node [draw=none, inner sep=0mm] (postC) at (\xpostC,\ymain) {
\begin{tikzpicture}[tdplot_main_coords, scale=0.5]
\cubenotikz{2}{1}{4}{1}{1}{2}{0.5}{black}{1}
\end{tikzpicture}
};

\node [aca] (acaT) at (\xacaT,\ymain) {\footnotesize \shortstack{$\ACA_{T}^2$}};

\node  [draw=none, above=0mm of acaT]  {
\scriptsize $\textcolor{red}{T} \times S \times D$};

\node [draw=none, inner sep=0mm] (cubeT) at (\xacaT,\ymain) {\begin{tikzpicture}[tdplot_main_coords, scale=0.5]
\cubenotikz{1}{1}{1}{1}{1}{1}{0.5}{black}{3}
\begin{pgfinterruptboundingbox}
\draw [draw=red, very thick, ->] (-\shift,0,0) -- (-\shift,0,\zmax);
\end{pgfinterruptboundingbox}
\end{tikzpicture}
};

\node [draw=none, inner sep=0mm] (postT) at (\xpostT,\ymain) {\begin{tikzpicture}[tdplot_main_coords, scale=0.5]
\cubenotikz{2}{1}{1}{1}{1}{1}{0.5}{black}{1}
\end{tikzpicture}
};

\node [draw=none] at (\xpostC,\ydim) {\scriptsize  $T \times S \times 2D$};

\node [aca, minimum height=\acaheight cm,text depth=\acaheight cm] (acaS) at (\xacaS,\ymain) {\footnotesize \shortstack{$\ACA_{S}^2$}};

\node  [draw=none, above=0mm of acaS]  {
\scriptsize $\textcolor{red}{S} \times D$};

\node [draw=none, inner sep=0mm] (cubeS) at (\xacaS,\ymain) {\begin{tikzpicture}[tdplot_main_coords, scale=0.5]
\cubenotikz{1}{1}{1}{1}{1}{1}{0.5}{black}{3}
\begin{pgfinterruptboundingbox}
\draw [draw=red, very thick, ->] (0,0,-\shift) -- (\xmax,0,-\shift);
\end{pgfinterruptboundingbox}
\end{tikzpicture}
};

\node [draw=none, inner sep=0mm] (postS) at (\xpostS,\ymain) {\begin{tikzpicture}[tdplot_main_coords, scale=0.5]
\cubenotikz{1}{1}{1}{1}{1}{1}{0.5}{black}{2}
\end{tikzpicture}
};

\node [draw=none] at (\xpostT,\ydim) {\scriptsize  $S \times 4D$};

\node [draw=none, below = 0mm of postS ] {\scriptsize $8D$};

\draw [very thick, ->] 
  (inputcube) -- (acaI) -- (postI) -- (acaC) -- (postC) -- (acaT) 
  -- (postT) -- (acaS) coordinate[midway] (skipbase) -- (postS);

\draw [very thick, ->] (skipbase) -- node[draw=none, left, pos=0.95]{\scriptsize \shortstack{sub-patch \\ skip \\ connection}}++(0,1);
\end{tikzpicture}
    }
    \caption{{\bf Universal Patch Encoder.}
    The input of this module is a patch with $C$ channels, $T$ time stamps, $S$ sub-patch of $I$ pixels, and a feature dimension of $D$.
    We use our proposed Axial Cross-Attention module (ACA) to sequentially collapse the pixel, channel, time, and sub-patch dimensions while simultaneously increasing the embedding dimension. This results in a single vector of dimension $8D$ and sub-patch embeddings of dimension $S\times 4D$.}
    \label{fig:upe}
\end{figure*}
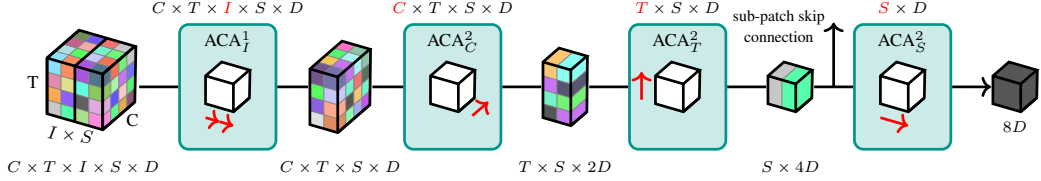

\subsection{\textsc{UniverSat}}
\label{sec:archi}

\begin{figure*}[t]
    \centering
    \resizebox{\linewidth}{!}{
    \input{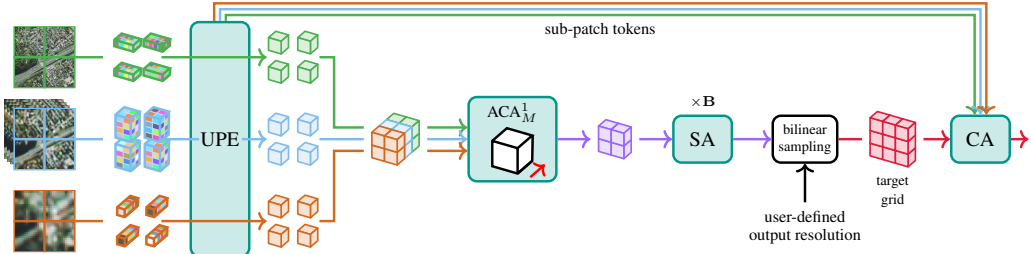}
    }
 \caption{\textbf{\textsc{UniverSat}.}
 A tile is observed by multiple sensors of arbitrary modality and resolution. Inputs are \emph{patchified} and embedded by a shared Universal Patch Encoder (UPE). The resulting tokens are stacked along the modality axis and collapsed to one token per patch via Axial Cross-Attention (ACA). We then apply $B$ self-attention (SA) blocks and resample the resulting feature map to the target resolution. Finally, the token attend high-resolution sub-token embeddings via cross-attention (CA) to recover fine spatial details.
}
    \label{fig:main}
\end{figure*}

We consider a tile partitioned into non-overlapping patches $\bP$.
Each patch $p \in \bP$ is observed by a subset of modalities $\bM$. In practice, $\bM$ may include multiple sensors within the same modality (\eg, Sentinel-2 and SPOT for optical imagery), and we use the term \emph{modality} to refer to these inputs generically. We denote by
$x_p^{m}$ the observation of patch $p$ under modality $m$.
Our objective is to produce a dense multimodal feature map
$\{z_p\}_{p \in \bP}$ at a \emph{user-specified} spatial resolution.
As illustrated in \cref{fig:main}, \textsc{UniverSat} follows a ViT-like backbone but differs in four key aspects:
(i) patches of arbitrary spatial, spectral, and temporal resolutions are embedded using a shared UPE,
(ii) co-registered modalities are fused via axial cross-attention,
(iii) fine spatial details are preserved through a sub-token skip connection,
(iv) the output resolution is specified at inference time.

\para{Patch-level modality fusion.}
For each patch $p$ and modality $m$, we compute
$
f_p^{m}, \; f_p^{m,\text{sub}} = \UPE(x_p^{m})~,
$
where $f_p^{m}$ is the patch embedding and
$f_p^{m,\text{sub}}$ are the corresponding sub-patch embeddings.
For a fixed patch $p$, the modality-specific embeddings
$\{f_p^{m}\}_{m\in\bM}$
are stacked along the modality axis and fused using an axial cross-attention module that collapses the modality dimension:
\begin{align}
f_p = \ACA^1_{M}\!\left(\{f_p^{m}\}_{m\in\bM}\right).
\end{align}
This yields a single multimodal embedding per patch.

\para{Spatial Transformer.} The multimodal patch embeddings
$f_\bP=\{f_p\}_{p\in\bP}$
are then processed by $B$ gated Transformer blocks~\citep{qiugated},
denoted $\SA^1, \cdots, \SA^B$:
\begin{align}
g_\bP = \SA^{B}\!\circ\cdots\circ\SA^{1}\!\big(f_\bP\big)~,
\end{align}
yielding refined patch embeddings $g$.
We use RoPE positional encodings over patch centers, scaled by patch size~\citep{reed2023scale}, and include four register tokens~\citep{darcet2024vision}.

\para{Any-Resolution Prediction.}
We can specify a target ground sampling distance (GSD), defining a new output spatial grid $\bQ$.
The patch embeddings are resampled to this grid via bilinear interpolation:
\begin{align}
g_{\bQ} = \mathrm{Bilinear}\big(g_\bP,\,\bP\!\to\!\bQ\big).
\end{align}
To recover fine spatial details lost during patch-level aggregation,
each target token attends to all sub-patch embeddings via cross-attention $\mathrm{CA}$ with a residual connection:
\begin{align}
z_{\bQ}
=
g_{\bQ}+
\mathrm{CA}
\left(
g_{\bQ},
\{f^{m,\text{sub}}_p\}_{p\in\bP,\,m\in\bM}
\right)~.
\end{align}
\subsection{Training}
\label{sec:training}

We exploit spatial alignment across modalities to train \textsc{UniverSat} in a self-supervised manner on multiple heterogeneous datasets.
Our objective combines (i) cross-modal contrast at the patch level and (ii) latent multimodal masked modeling (LM$_3$).

\para{Scale Augmentation and Masking.}
We train with aggressive input dropping to improve robustness across scales and sensor configurations.
For each tile, we sample the input patch size and target output resolution from dataset-specific presets, then apply a compositional masking operator that drops modalities, timestamps, channels, and patches.
This produces a visible subset of modalities $\bM^\star$ and patches $\bP^\star$, while masked patches $\bD=\bP\setminus\bP^\star$ are used for masked prediction.
Overall, approximately $90\%$ of input atoms are removed, reducing computation and encouraging invariance across modalities, temporal sampling, spectral density, and spatial scale.
Sampling details are given in the Appendix.

\para{Cross-modal contrast.}
Different modalities observing the same patch share a common latent variable: their content~\citep{astruc2024omnisat,tseng2022croco}.
We encourage modality-invariant representations by applying a batch-wise, multi-positive contrastive loss $\mathcal{L}_{\text{con}}$~\citep{astruc2024anysat} to the UPE embeddings $\{f^m_p\}_{p \in \bP^\star,\, m \in \bM^\star}$ of visible patches.
Embeddings of the same patch in different modalities are positives, embeddings of different patches in different modalities are negatives, and pairs from different patches in the same modality are excluded, as they are trivially separable.

\para{Linear Multimodal Masked Modeling (LM$_3$).}
We extend latent masked image modeling~\citep{wei2024towards} to multimodal and multitemporal EO data by predicting representations of masked patches at selected timestamps.
Following~\citep{herzog2025olmoearth}, we define targets in a latent space using frozen random projections.
Concretely, for each modality $m$, we use a randomly initialized linear projection $\phi^{\text{rand}}_m$ that maps monotemporal input patches to a fixed-size target space and remains frozen during training.

A key challenge is to balance spatial and temporal supervision.
Using all timestamps may lead the model to exploit trivial temporal cues, while ignoring time degrades representation quality.
We therefore sample a small set of timestamps per tile and assign each masked patch $p \in \bD$ a target time $\tau(p)$.
Given the visible multimodal patch embeddings $\{z_q\}_{q \in \bP^\star}$, a predictor $\phi^{\text{pred}}$ infers representations for masked patches conditioned on $\tau(p)$:
\begin{align}
\left\{z_p^{\tau(p)}\right\}_{p \in \bD}
=
\phi^{\text{pred}}\big(\{z_q\}_{q \in \bP^\star}\big).
\end{align}

Modality-specific heads $\phi^{\text{head}}_m$ then map these embeddings to the target space, supervised with a contrastive objective in which the target of the same patch, modality, and timestamp is the only positive, and the targets of all other masked patches are negatives:
\begin{align}
\mathcal{L}_{\text{LM}^3}
=
\sum_{m \in \bM}
\sum_{p \in \bD}
\text{InfoNCE}\Big(
\phi^{\text{head}}_m\big(z_p^{\tau(p)}\big),
\phi^{\text{rand}}_m\big(x_p^m[\tau(p)]\big)
\Big).
\end{align}
Although the prediction heads $\phi^{\text{head}}_m$ are modality-specific, they all take the same multimodal patch representation $z_p$ as input. Solving LM$_3$ therefore requires $z_p$ to preserve information from all modalities in a shared latent space. As the modality-specific heads are used only during pretraining, the model itself remains modality agnostic. Finally, since the targets are random projections and not learned, the objective avoids collapse. See Appendix for more details.
Random projections give targets of the same size for sensors whose raw patches differ in dimension by orders of magnitude, and InfoNCE is less sensitive than a regression loss to the widely varying scale of these targets; both choices clearly outperform their alternatives (\cref{tab:ablation}).
The two losses are complementary: $\mathcal{L}_{\text{con}}$ only supervises the UPE and aligns the content shared across modalities, whereas $\mathcal{L}_{\text{LM}^3}$ trains the full encoder and requires $z_p$ to retain modality-specific information.

\begin{figure*}[t]%
    \centering
    \input{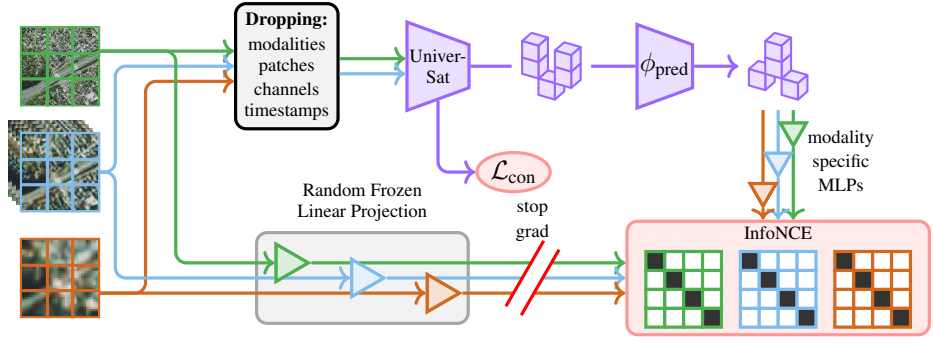}
\caption{\textbf{Training Scheme.}
We give a \textsc{UniverSat} network a heavily masked version of the input patches. We apply a cross-modal contrastive loss to harmonize the output of its Universal Patch Encoders.
We then give the embedded patches that were not masked to a decoder network, which tries to predict the values of random projections of the raw input patches. We use a batch-wise InfoNCE loss on each modality separately.
}
    \label{fig:training}
\end{figure*}

\para{Final Loss.}
The final loss writes:
$
    \mathcal{L}\!=\!
    \mathcal{L}_{\text{LM}^3}
    \!+\!
    \lambda_\text{con} \mathcal{L}_\text{con}
$
, where $\lambda_\text{con}$ is a non-negative hyperparameter.

\section{Experiments}
\label{sec:Benchmark}
\vspace{-0mm}

We first describe the experimental setting and our multi-dataset training corpus (\cref{sec:exp:setting}).
We then report probing experiments demonstrating how the embeddings learned by \textsc{UniverSat} transfer to diverse and challenging EO tasks (\cref{sec:exp:eval}).
Finally, we provide an ablation study analyzing the impact of key architectural and training choices (\cref{sec:ablation}).

\subsection{Experimental setting}
\label{sec:exp:setting}

\paragraph{Training Set.}
We train a single \textsc{UniverSat} model on a composite corpus of seven heterogeneous datasets:
FLAIR-Hub~\citep{ign2025flairhub},
PASTIS-HD~\citep{garnot2021panoptic,astruc2024omnisat},
TSAI-TS~\citep{ahlswede2022treesatai,astruc2024omnisat},
Planted~\citep{pazos2024planted},
S2NAIP-Urban~\citep{bastani2023satlaspretrain,satlassuperres},
HyperGlobal~\citep{hypersigma},
and the NEON subset of EarthView~\citep{velazquez2025earthview}.
Note that folds~4 and~5 of PASTIS, used for validation and testing in our benchmarks, are excluded from the unsupervised training set, even though no labels from PASTIS are used during pretraining.
An overview is provided in \cref{tab:dataset}, with full details in the appendix.
The sensors span a wide range of acquisition conditions: spatial resolutions from $0.1$ to $300$\,m, temporal depth from $1$ to $140$ images per year, and spectral width from $1$ to $396$ channels, with tile extents ranging from $0.4$ to $600$\,ha.
To our knowledge, \textsc{UniverSat} is the first EO foundation model jointly trained on such a broad set of $13$ sensors and modalities, including some that are rarely incorporated into large-scale pretraining, such as very-high-resolution hyperspectral time series.

\subsection{Benchmarks}
\vspace{-0mm}
\label{sec:exp:eval}
We evaluate our model on 16 datasets from GeoBench~\citep{lacoste2023geo}, PangaeaBench~\citep{marsocci2024pangaea}, and SpectralEarth~\citep{braham2025spectralearth}.
\textsc{UniverSat} remains competitive with less versatile approaches in standard settings, while also delivering strong performance on more specialized tasks such as hyperspectral analysis.

\paragraph{Probing Experiments.}
In \cref{tab:probing}, we report results in a strict probing setting (kNN and linear probing only), following the protocol of~\citet{tseng2025galileo}.
Our model achieves strong performance across a wide range of datasets, including state-of-the-art results on BrickKiln~\citep{lee2021scalable} and Sen1Flood11~\citep{bonafilia2020sen1floods11}, despite being significantly more general than competing approaches, which are typically restricted to specific sensors (\eg, Sentinel-1/2 or Landsat) or single timestamps.

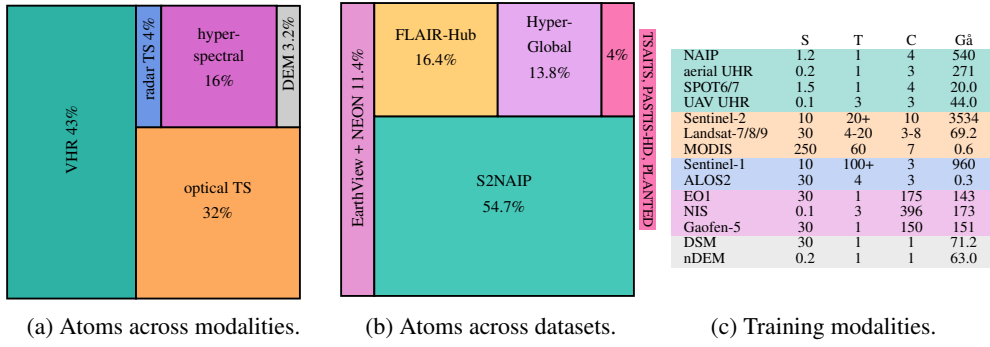
\begin{figure*}[t]
    \centering
    \definecolor{colCop}{RGB}{255,208,120}%
\definecolor{colSpec}{RGB}{230,150,210}%
\definecolor{ColS2N}{RGB}{70,200,185}%
\definecolor{ColFla}{RGB}{255,160,135}%
\definecolor{ColHyp}{RGB}{230,170,240}%
\definecolor{ColOther}{RGB}{250,120,180}%
\def\printonlylargeenough#1#2{
\unless\ifdim#2pt<#1pt\relax #2\printnumbertrue
\else \printnumberfalse \fi}
\newif\ifprintnumber 
\definecolor{colVHR}{RGB}{46,178,164}%
\definecolor{colOpt}{RGB}{254,170,94}%
\definecolor{colRadar}{RGB}{110,150,230}%
\definecolor{colHyper}{RGB}{214,106,205}%
\definecolor{colDEM}{RGB}{205,205,205}%
\begin{tabular}{c@{}c@{}c}
\begin{subfigure}{0.31\textwidth}
\resizebox{!}{.9\linewidth}{
\begin{tikzpicture}[yscale=1.0]
  \pie[
    radius=2.2,
    square,
    sum=auto,
    text=inside,
    line width=0.1pt,
    before number=\printonlylargeenough{100}, after number=\ifprintnumber\%\fi,
    color={colVHR, colOpt, colRadar, colHyper, colDEM},
    every node/.append style={transform shape=false}
  ]{
    43/\rotatebox{90}{\scriptsize VHR 43\%},
    32/{\scriptsize optical TS \\ \scriptsize 32\%},
    3.5/{\scriptsize\rotatebox{90}{\!\!\!\!radar TS 4\%}},
    16/{\scriptsize  \shortstack{~\\[0mm]hyper-\\spectral} \\[0mm] \scriptsize 16\%}, %
    3.2/\rotatebox{+90}{\raisebox{0mm}{\scriptsize DEM 3.2\%}}
  }
\end{tikzpicture}}
\caption{Atoms across modalities.}
\label{fig:datasets:modal}
\end{subfigure}
&
\begin{subfigure}{0.31\textwidth}
\raisebox{1.5mm}{
\resizebox{!}{.9\linewidth}{
\begin{tikzpicture}[yscale=1.0]
  \pie[
    radius=2.2,
    square,
    sum=auto,
    text=inside,
    line width=0.1pt,
    before number=\printonlylargeenough{100}, after number=\ifprintnumber\%\fi,
    color={colSpec, ColS2N, colCop, ColHyp, ColOther}, 
    every node/.append style={transform shape=false}
  ]{
    11.4/{\\[0mm]\scriptsize \rotatebox{90}{EarthView + NEON 11.4\%}},
    54.7/{\scriptsize S2NAIP\\ \scriptsize 54.7\%},
    16.4/{\scriptsize FLAIR-Hub \\ \scriptsize 16.4\%}, %
    13.8/{\scriptsize Hyper- \\ \scriptsize Global \\\scriptsize 13.8\% }, %
    4.3/{\\[0mm]\scriptsize 4\%\quad}%
  }
 \node at (2.4,+0.3) {\rotatebox{-90}{\colorbox{ColOther}{\tiny TSAITS, PASTIS-HD, PLANTED}}};
\end{tikzpicture}}}
\vspace{-5mm}
\caption{Atoms across datasets.}
\label{fig:datasets:sensors2}
\end{subfigure}
&

\begin{subfigure}{0.32\textwidth}
\raisebox{5mm}{
\resizebox{\linewidth}{!}{
\scriptsize{
\begin{tabular}{lcccc}
    & S & T & C & G\aa \\
     \rowcolor{colVHR!40}{\vphantom{/}NAIP} & 1.2 & 1 & 4 & 540 \\[0mm]
     \rowcolor{colVHR!40}{\vphantom{/}aerial UHR} & 0.2 & 1 & 3 & 271\\[0mm]
     \rowcolor{colVHR!40}{\vphantom{/}SPOT6/7} & 1.5 & 1 & 4 & 20.0\\[0mm]
     \rowcolor{colVHR!40}{\vphantom{/}UAV UHR} & 0.1 & 3 & 3 & 44.0\\[0mm]
     \rowcolor{colOpt!40}{\vphantom{/}Sentinel-2} & 10 & 20+ & 10 & 3534\\[0mm]
     \rowcolor{colOpt!40}{\vphantom{/}Landsat-7/8/9} & 30 & 4-20 & 3-8 & 69.2\\[0mm]
     \rowcolor{colOpt!40}{\vphantom{/}MODIS}& 250 & 60& 7 & 0.6\\[0mm]
     \rowcolor{colRadar!40}{\vphantom{/}Sentinel-1} & 10 & 100+ & 3 & 960\\[0mm]
     \rowcolor{colRadar!40}{\vphantom{/}ALOS2} & 30 & 4 & 3 & 0.3 \\[0mm]
     \rowcolor{colHyper!40}{\vphantom{/}EO1} & 30 & 1 & 175 & 143 \\[0mm]
     \rowcolor{colHyper!40}{\vphantom{/}NIS} & 0.1 & 3 & 396 & 173\\[0mm]
     \rowcolor{colHyper!40}{\vphantom{/}Gaofen-5} & 30 & 1 & 150 & 151\\[0mm]
     \rowcolor{colDEM!40}{\vphantom{/}DSM}& 30 & 1 & 1 & 71.2\\[0mm]
     \rowcolor{colDEM!40}{\vphantom{/}nDEM} & 0.2 & 1 & 1 & 63.0\\
\end{tabular}
}
}
}
~\\
\caption{Training modalities.}
\label{fig:datasets:sensors}
\end{subfigure}

\end{tabular}

\caption{\textbf{Training Datasets.}
Distribution of atoms (\aa), defined as one pixel, one band, and one timestamp~\citep{de2025atomizer}, across modalities (\cref{fig:datasets:modal}) and datasets (\cref{fig:datasets:sensors2}).
\Cref{fig:datasets:sensors} summarizes the supported sensors, reporting their typical spatial resolution ($S$, in meters), temporal depth ($T$, in images per year), number of channels ($C$), and total number of atoms (G\aa).}
    \label{tab:dataset}
\end{figure*}

\begin{table}[b]
    \centering
    \resizebox{1\linewidth}{!}{
{\scriptsize
\begin{tabular}{
l@{\;\;}l
x{9mm} y{9mm} x{9mm}
y{11mm} x{12mm} y{10mm} x{12mm}
}
&&
\multicolumn{3}{c}{classification}
&
\multicolumn{4}{c}{semantic segmentation}
\\ \cmidrule(lr){3-5} \cmidrule(lr){6-9}
\multicolumn{2}{c}{
\shortstack{
\adjustbox{cfbox=blue 1pt 0pt}{unseen config}
\\
\adjustbox{cfbox=red 1pt 0pt}{unseen sensor}
}
}
&
\rotatebox{0}{\shortstack{m-brick-\\kiln}} &
\rotatebox{0}{\shortstack{m-\\pv4ger}} &
\rotatebox{0}{\shortstack{m-forest\\net}} &
\rotatebox{0}{\shortstack{PASTIS\\-R}} &
\rotatebox{0}{\shortstack{Sen1\\Floods11}} &
\rotatebox{0}{\shortstack{m-chesa\\peake}} &
\rotatebox{0}{\shortstack{m-Neon\\Tree}}\\

\multicolumn{2}{l}{\makecell[l]{Modality}} &
\adjustbox{cfbox=blue 1pt 0pt}{\makecell{S2\mono}} &
\adjustbox{cfbox=blue 1pt 0pt}{S2\mono} &
\adjustbox{cfbox=blue 1pt 0pt}{L8\mono} &
\makecell{S1+S2\hphantom{+}} &
\adjustbox{cfbox=blue 1pt 0pt}{S1\mono} &
aerial&
\adjustbox{cfbox=red 1pt 0pt}{RGB aerial}
\\

\multicolumn{2}{l}{Metric} &
Acc. & Acc. & Acc. & mIoU &
mIoU& mIoU& mIoU\\
\midrule
\multirow{9}{*}{\rotatebox{90}{SSL}}

& DINOv2 \citep{oquab2023dinov2} & - & - & - & - & - & \rankcell{64.0}{2} & \rankcell{59.1}{1}\\

& DINOv3 7B \citep{simeoni2025dinov3} & 91.3 & - & \rankcell{48.0}{2} & - & - & - & -\\

& CROMA-L \citep{fuller2023croma} & 91.7 & 94.5 & - & 44.4 & 78.8 & - & -\\

& CopernicusFM \citep{wang2025unifiedcopernicusfoundationmodel} & 85.9 & - & - & - & 77.6 & 32.2 & -\\

& AnySat \citep{astruc2024anysat} & 84.5 & 90.3 & 34.0 & \rankcell{46.2}{3} & 77.8 & \rankcell{61.7}{3} & 49.9\\

& DOFA \citep{xiong2024dofa} & - & \rankcell{95.8}{2} & - & 13.4 & 77.4 & 59.2 & \rankcell{55.4}{2}\\

& Panopticon \citep{waldmann2025panopticon} & \rankcell{92.9}{3} & \rankcell{96.7}{1} & \rankcell{52.3}{1]} & - & - & 60.8 & 50.4\\

& Satlas \citep{bastani2023satlaspretrain} & 83.0 & - & 36.9 & 28.0 & 72.9 & - & -\\

& \bf UniverSat-B (ours) & \rankcell{94.5}{1} & \rankcell{92.7}{3} & \rankcell{41.7}{3} & \rankcell{47.9}{2} & \rankcell{80.1}{1}& \rankcell{64.7}{1} & \rankcell{51.0}{3}\\

\greyrule
\multirow{3}{*}{\rotatebox{90}{SSup}}
& Galileo-B \citep{tseng2025galileo} & 91.1 & 93.1 & - & 39.2 & \rankcell{79.4}{3} & - & -\\

& TerraMind \citep{jakubik2025terramind} & 91.9 & - & - & - & 78.7 & - & -\\

& OlmoEarth-L \citep{herzog2025olmoearth} & \rankcell{93.4}{2} & - & 41.6 & \rankcell{51.8}{1} &  \rankcell{79.8}{2} & - & -\\
\bottomrule
\end{tabular}
}}

\if 10
\resizebox{\linewidth}{!}{
{\scriptsize
\begin{tabular}{
l@{\,}l
x{9mm} y{9mm} x{9mm}
y{11mm} x{12mm} y{10mm} x{10mm}
}
&&
\multicolumn{3}{c}{classification}
&
\multicolumn{4}{c}{semantic segmentation}
\\ \cmidrule(lr){3-5} \cmidrule(lr){6-9}
&&
\rotatebox{0}{\shortstack{brick-\\kiln}} &
\rotatebox{0}{pv4ger} &
\rotatebox{0}{\shortstack{forest\\net}} &
\rotatebox{0}{\shortstack{PASTIS\\-R}} &
\rotatebox{0}{\shortstack{Sen1\\Floods11}} &
\rotatebox{0}{\shortstack{chesa\\peake}} &
\rotatebox{0}{\shortstack{SA-\\crop}}\\

\multicolumn{2}{l}{\makecell[l]{Modality}} &
\adjustbox{cfbox=blue 1pt 0pt}{\makecell{S2\mono}} &
\adjustbox{cfbox=blue 1pt 0pt}{S2\mono} &
\adjustbox{cfbox=blue 1pt 0pt}{L8\mono} &
\makecell{S1+S2\hphantom{+}} &
\adjustbox{cfbox=blue 1pt 0pt}{S1\mono} &
aerial &
aerial
\\

\multicolumn{2}{l}{Metric} &
Acc. & Acc. & Acc. & mIoU &
mIoU& mIoU& mIoU\\

\midrule
\multirow{9}{*}{\rotatebox{90}{SSL}}

& DINOv2 \citep{oquab2023dinov2} & - & - & - & - & - & \rankcell{64.0}{1} & 32.0 \\

& DINOv3 7B \citep{simeoni2025dinov3} & 91.3 & - & \rankcell{48.0}{2} & - & - & - & -\\

& CROMA-L \citep{fuller2023croma} & 91.7 & 94.5 & - & 44.4 & 78.8 & - & \rankcell{34.4}{2}\\

& CopernicusFM \citep{wang2025unifiedcopernicusfoundationmodel} & 85.9 & - & - & - & 77.6 & 32.2 & -\\

& AnySat \citep{astruc2024anysat} & 84.5 & 90.3 & 34.0 & \rankcell{46.2}{3} & 77.8 & \rankcell{61.7}{3} & 27.2\\

& DOFA \citep{xiong2024dofa} & - & \rankcell{95.8}{2} & - & 13.4 & 77.4 & 59.2 & \rankcell{32.5}{3}\\

& Panopticon \citep{waldmann2025panopticon} & \rankcell{92.9}{3} & \rankcell{96.7}{1} & \rankcell{52.3}{1]} & - & - & 60.8 & \rankcell{35.7}{1}\\

& Satlas \citep{bastani2023satlaspretrain} & 83.0 & - & - & 28.0 & 72.9 & - & 27.1\\

& \bf UniverSat-B (ours) & \rankcell{93.5}{1} & \rankcell{95.1}{3} & \rankcell{41.0}{3} & \rankcell{46.9}{2} & \rankcell{80.8}{1}& \rankcell{62.0}{2} & 27.7\\

\greyrule
\multirow{3}{*}{\rotatebox{90}{SSup}}
& Galileo-B \citep{tseng2025galileo} & 91.1 & 93.1 & - & 39.2 & \rankcell{79.4}{3} & - & - \\

& TerraMind \citep{jakubik2025terramind} & 91.9 & - & - & - & 78.7 & - & - \\

& OlmoEarth-L \citep{herzog2025olmoearth} & \rankcell{93.4}{2} & - & - & \rankcell{51.8}{1} &  \rankcell{79.8}{2} & - & - \\
\bottomrule
\end{tabular}
}}

\fi

    \caption{{\bf Probing Experiment.} We evaluate our model on several classification and segmentation datasets using kNN for classification and linear probing for segmentation. SSL designates methods trained with only sensor observation, and SSup models that use labels as supervision or training modalities. Results of competing methods are taken from the literature \citep{tseng2025galileo,waldmann2025panopticon,herzog2025olmoearth}. When values conflict, we take the highest. Top 3 performance are \colorbox{green!20}{highlighted}. \scriptsize{\mono}\normalsize : single time stamp.}
    \label{tab:probing}
\end{table}

On PangaeaBench (\cref{tab:pangaea}), unlike prior work using large decoder heads (\eg, UperNet~\citep{xiao2018unified}), we exploit \textsc{UniverSat}'s dense embeddings and perform semantic segmentation with a simple linear probe.
Despite using $3700$--$5000\times$ fewer supervised parameters, our model remains competitive and reaches state-of-the-art performance on PASTIS-R~\citep{garnot2021panoptic} and AI4Farms~\citep{persello2023ai4smallfarms}.
In particular, it outperforms RAMEN~\citep{houdre2025ramen}, which also uses a shared encoder but resamples inputs to a common format, on two of the four datasets.
Under full end-to-end fine-tuning, \textsc{UniverSat} trains stably and reaches state-of-the-art performance on Sen1Floods11 (\cref{sec:app:finetuning}).

Importantly, we separate unseen \emph{configurations} of known sensor families---mono-temporal Sentinel-1/2, reduced band subsets, and derived products such as HLS---from genuinely \emph{unseen sensors}, absent from pretraining in any form: PlanetScope and EnMAP.
\textsc{UniverSat} is robust in both regimes: on DynamicEarthNet (PlanetScope), its $9$K linear probe reaches $37.1$ mIoU, above a fully supervised ViT ($36.8$) and close to UperNet decoders using $4400\times$ more supervised parameters.

\begin{table}[t]
    \centering
    \begin{minipage}{0.35\linewidth}
    \caption{{\bf Probing with Decoders.} We evaluate our model on datasets from the Pangaea Benchmark using a linear probe, whereas other models use a heavyweight UperNet decoder. ViT-B \emph{sup.} is a fully supervised baseline trained end-to-end~\citep{marsocci2024pangaea}. \scriptsize{\mono}\normalsize : single time stamp. \protect\adjustbox{cfbox=blue 1pt 0pt}{unseen config.}, \protect\adjustbox{cfbox=red 1pt 0pt}{unseen sensor}.}
    \label{tab:pangaea}
    \end{minipage}
    \hspace{1mm}
    \begin{minipage}{0.63\linewidth}
    \resizebox{\linewidth}{!}{
    \resizebox{\linewidth}{!}{
\begin{tabular}{ll x{13mm}  y{10mm} x{13mm} y{15mm}}
     & \multirow{2}{*}{{\rotatebox{0}{\shortstack{Probe\\param}}}} & 
     \multicolumn{1}{c}{\rotatebox{0}{\shortstack{PASTIS-R}}} & 
     \multicolumn{1}{c}{\rotatebox{0}{\shortstack{BurnScar}}}&
     \multicolumn{1}{c}{\rotatebox{0}{\shortstack{DynEN}}}&
     \multicolumn{1}{c}{\rotatebox{0}{\shortstack{AI4Farms}}}\\
     \multicolumn{1}{l}{Modality}&&
     \multicolumn{1}{c}{S1+S2}& 
     \multicolumn{1}{c}{\adjustbox{cfbox=blue 1pt 0pt}{HLS\mono}}
     &
     \multicolumn{1}{c}{\adjustbox{cfbox=red 1pt 0pt}{PlanetScope}}
     &
     \multicolumn{1}{c}{\adjustbox{cfbox=blue 1pt 0pt}{S2\mono \tiny{RGBNiR}}}
     \\[-1mm] \midrule
    ViT-B \emph{sup.} \citep{marsocci2024pangaea} & 125M & 38.5 & 81.6 & 36.8 & \rankcell{38.4}{3} \\\greyrule
    Satlas \citep{bastani2023satlaspretrain} & 33M & 17.5 & 80.0& 36.3 & 25.1 \\
    CROMA \citep{fuller2023croma} & 47M & 32.3 & 82.4 & \rankcell{38.3}{3} & 25.7 \\
    DOFA + \citep{xiong2024dofa} & 47M & 40.3 & \rankcell{86.5}{1}& \rankcell{41.1}{1} & 29.9 \\
    Ramen \citep{houdre2025ramen} & 40M & \rankcell{42.3}{3} & \rankcell{85.0}{2}  & \rankcell{39.9}{2} & \rankcell{38.8}{2}  \\
   TerraMind \citep{jakubik2025terramind} & 47M & \rankcell{43.1}{2} & \rankcell{82.9}{3} & 37.9 & 27.5 \\\greyrule
    \bf UniverSat-B & \hphantom{\,1}9K & \rankcell{47.9}{1} & 81.5 & 37.1 & \rankcell{41.1}{1}\\
    \bottomrule
\end{tabular}
}

\if 10

\resizebox{\linewidth}{!}{
\begin{tabular}{ll x{13mm}  y{10mm} x{10mm} y{15mm}}
     & \multirow{2}{*}{{\rotatebox{0}{\shortstack{Probe\\param}}}} & 
     \multicolumn{1}{c}{\rotatebox{0}{\shortstack{PASTIS-R}}} & 
     \multicolumn{1}{c}{\rotatebox{0}{\shortstack{BurnScar}}} &
     \multicolumn{1}{c}{\rotatebox{0}{SN7}} &
     \multicolumn{1}{c}{\rotatebox{0}{\shortstack{AI4Farms}}}\\
     \multicolumn{1}{l}{Modality}&&
     \multicolumn{1}{c}{S1+S2}& 
     \multicolumn{1}{c}{\adjustbox{cfbox=blue 1pt 0pt}{HLS\mono}} &
     \multicolumn{1}{c}{aerial}
     &
     \multicolumn{1}{c}{\adjustbox{cfbox=blue 1pt 0pt}{S2\mono \tiny{RGBNiR}}}
     \\[-1mm] \midrule
    Satlas \citep{bastani2023satlaspretrain} & 33M & 17.5 & 80.0 & 61.9 & 25.1 \\
    CROMA \citep{fuller2023croma} & 47M & 32.3 & 82.4 & 58.3 & 25.7 \\
    DOFA + \citep{xiong2024dofa} & 47M & 40.3 & \rankcell{86.5}{1} & 63.1 & \rankcell{29.9}{3} \\
    Ramen \citep{houdre2025ramen} & 40M & \rankcell{42.3}{3} & \rankcell{85.0}{2}  & 60.3 & \rankcell{38.8}{2}  \\
   TerraMind \citep{jakubik2025terramind} & 47M & \rankcell{43.1}{2} & \rankcell{82.9}{3} & 60.0 & 27.5 \\\greyrule
    \bf UniverSat-B & \hphantom{\,1}9K & \rankcell{46.9}{1} & 81.2 & - & \rankcell{41.0}{1}\\
    \bottomrule
\end{tabular}
}

\fi

\if10
\resizebox{\linewidth}{!}{
\begin{tabular}{ll x{10mm}  y{10mm} x{10mm} y{10mm}  x{10mm} y{10mm} x{10mm}}
     & \multirow{2}{*}{{\rotatebox{0}{\shortstack{Probe\\param}}}} & 
     \multicolumn{1}{c}{\rotatebox{0}{\shortstack{PASTIS\\R}}} & 
     \multicolumn{1}{c}{\rotatebox{0}{\shortstack{Burn\\Scar}}} & 
     \multicolumn{1}{c}{\rotatebox{0}{MADOS}} &
     \multicolumn{1}{c}{\rotatebox{0}{Sen1F11}}  &
     \multicolumn{1}{c}{\rotatebox{0}{SN7}} &
     \multicolumn{1}{c}{\rotatebox{0}{\shortstack{AI4\\Farms}}}\\
     \multicolumn{1}{l}{Modality}&&
     \multicolumn{1}{c}{S1+S2}& 
     \multicolumn{1}{c}{\adjustbox{cfbox=blue 1pt 0pt}{HLS\mono}} &
     \multicolumn{1}{c}{\adjustbox{cfbox=blue 1pt 0pt}{S2\mono}} &
     \multicolumn{1}{c}{\adjustbox{cfbox=blue 1pt 0pt}{S1 \mono}} &
     \multicolumn{1}{c}{aerial}
     \\[-1mm] \midrule
    Satlas \citep{bastani2023satlaspretrain} & 33M & 17.5 & 80.0 & 55.9 & \rankcell{90.3}{3} & 61.9 & 25.1 \\
    CROMA \citep{fuller2023croma} & 47M & 32.3 & 82.4 & 67.6 & 89.9 & 58.3 & 25.7 \\
    DOFA + \citep{xiong2024dofa} & 47M & 40.3 & \rankcell{86.5}{1} & \rankcell{69.8}{3} & \rankcell{90.4}{2}  & 63.1 & 29.9 \\
    Ramen \citep{houdre2025ramen} & 40M & \rankcell{42.3}{3} & \rankcell{85.0}{2} & \rankcell{69.7}{3} & \rankcell{91.0}{1} & 60.3 & 38.8  \\
   TerraMind \citep{jakubik2025terramind} & 47M & \rankcell{43.1}{2} & \rankcell{82.9}{3} & \rankcell{75.6}{1} & 90.1 & 60.0 & 27.5 \\\greyrule
    \bf UniverSat-SSL & 9K & \rankcell{46.9}{1} & 81.2 & 45.8 & 80.8 & - & 41.0\\
    \bottomrule
\end{tabular}
}
\fi

    }
    \end{minipage}
\end{table}

\begin{table}[b]
    \centering
    \resizebox{\linewidth}{!}{
\scriptsize
\begin{tabular}{
@{\,}l cc
y{6mm}  x{6mm} y{7mm} x{6mm} y{7mm} x{6mm} y{11mm}
}
\multirow{3}{*}[-7mm]{
\shortstack{\adjustbox{cfbox=red 1pt 0pt}{unseen sensor}  \\all datasets use\\\adjustbox{cfbox=red 1pt 0pt}{EnMAP}}
}
&&
&
\multicolumn{6}{c}{Segmentation (mIoU)}&
Class. (F1)
\\\cmidrule(lr){4-9}\cmidrule(lr){10-10}
&
\multirow{2}{*}{\rotatebox{90}{\shortstack{Free\\param.}}} &
\multirow{2}{*}[3.5mm]{\raisebox{0mm}{\rotatebox{90}{\shortstack{Trained w.\\EnMAP}}}} &
\rotatebox{60}{Cdl} &
\rotatebox{60}{Nlcd}&
\rotatebox{60}{\shortstack{Euro\\crops}} &
\rotatebox{60}{\shortstack{Tree\\map}} &
\rotatebox{60}{\shortstack{BD\\Foret}} &
\rotatebox{60}{BnetD} &
\rotatebox{60}{Corine} \\
&
&
&
\crop&
\landcover&
\crop&
\tree&
\tree&
\tree&
\landcover
\\
\midrule
\textcolor{black!50}{SpectralEarth-L \citep{braham2025spectralearth}} 
& \textcolor{black!50}{1.2M}
& \textcolor{black!50}{\yes}
&
\textcolor{black!50}{72.7} & \textcolor{black!50}{42.7} & \textcolor{black!50}{62.1} & \textcolor{black!50}{42.6} & \textcolor{black!50}{71.1} & \textcolor{black!50}{44.0} & \textcolor{black!50}{74.0}
\\
\textcolor{black!50}{SpatSIGMA-B \citep{hypersigma}} & \textcolor{black!50}{900K}& & \textcolor{black!50}{63.5} & \textcolor{black!50}{36.0} &\textcolor{black!50}{50.7} & \textcolor{black!50}{37.0} & \textcolor{black!50}{58.2} &  \textcolor{black!50}{38.6} & \textcolor{black!50}{68.2}\\\midrule
DOFA-L \citep{xiong2024dofa} & 1.2M& \yes & 60.6 & 33.4 & 46.1 & 34.7 &  51.3 &  37.4 & 67.5 \\
\bf UniverSat SSL & 900K & & \bf 64.7 & \bf 39.0 & \bf 55.3 & \bf 38.2 & \bf 63.3 & \bf 43.9 & \bf 73.6 \\
\bf UniverSat SSL w. linear & \hphantom{99}9K & & 61.4 & 34.6 & 52.8 & 35.3 & 57.5 & 40.3 & 71.5\\
\bottomrule
\end{tabular}
}

\caption{{\bf Hyperspectral performance.}
We evaluate \textsc{UniverSat} on the SpectralEarth benchmark, composed of multiple tasks using EnMAP hyperspectral data.
We compare against specialized hyperspectral models (top) and general foundation models supporting hyperspectral inputs (bottom).
Despite no training on EnMAP, \textsc{UniverSat} surpasses the state-of-the-art DOFA foundation model, and remains competitive even when using only a linear probe instead of a convolutional decoder.}
    \label{tab:hyper}
\end{table}

\paragraph{Hyperspectral Data.}
We evaluate our model on the SpectralEarth benchmark~\citep{braham2025spectralearth} (\cref{tab:hyper}), which consists of multiple tasks based on EnMAP hyperspectral imagery (up to 500 bands), following their probing protocol.
Compared to DOFA~\citep{xiong2024dofa}, a foundation model trained on EnMAP, our approach achieves consistently better performance across all tasks, despite not being trained on EnMAP.

We also compare to specialized hyperspectral models.
Our method outperforms SpatSIGMA~\citep{hypersigma} and approaches the performance of SpectralEarth-L, a model specifically designed for EnMAP and trained with self-supervision on the evaluation data itself.
These results highlight that, beyond its versatility, our model remains highly competitive in specialized regimes such as hyperspectral analysis.

\begin{figure}
    \centering
    \begin{tabular}{@{}cccc@{}}
\includegraphics[width=0.21\linewidth, height=0.21\linewidth]{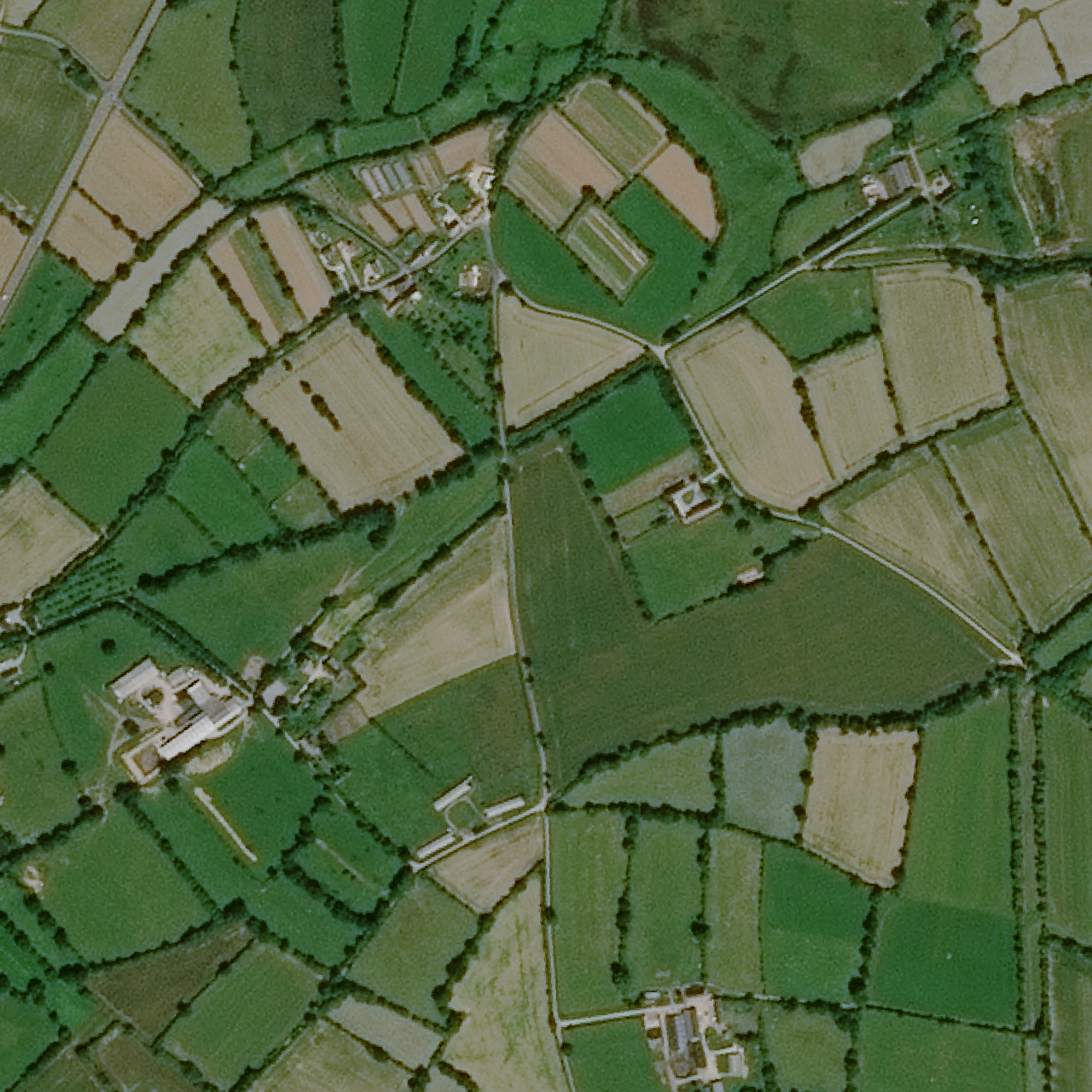}
&
\includegraphics[width=0.21\linewidth, height=0.21\linewidth]{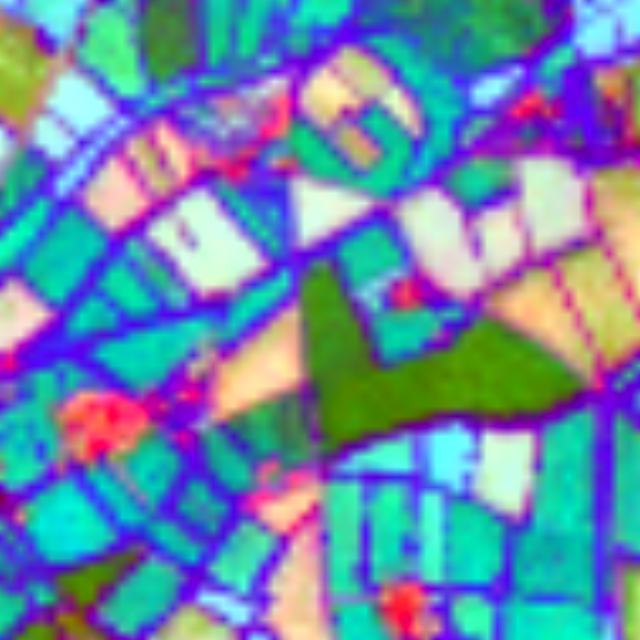}
&
\includegraphics[width=0.21\linewidth, height=0.21\linewidth]{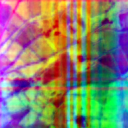}
&
\includegraphics[width=0.21\linewidth, height=0.21\linewidth]{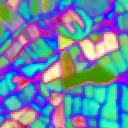}
\\
Spot6 (1.5m) & \textsc{UniverSat} (2m) & OlmoEarth (10m) & AlphaEarth (10m) \\
\raisebox{13.0mm}{
\begin{tabular}{@{}c@{\,}c@{\,}c@{}}
\includegraphics[width=0.065\linewidth, height=0.065\linewidth]{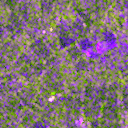}&
\includegraphics[width=0.065\linewidth, height=0.065\linewidth]{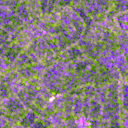}&
\includegraphics[width=0.065\linewidth, height=0.065\linewidth]{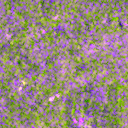}\\
\multicolumn{3}{c}{ S1 T.S (10m)}\\
\includegraphics[width=0.065\linewidth, height=0.065\linewidth]{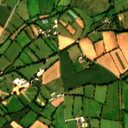}&
\includegraphics[width=0.065\linewidth, height=0.065\linewidth]{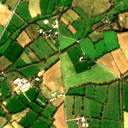}&
\includegraphics[width=0.065\linewidth, height=0.065\linewidth]{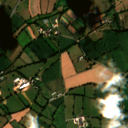}\\
\multicolumn{3}{c}{S2 T.S. (10m)}
\end{tabular}
}
&
\includegraphics[width=0.21\linewidth, height=0.21\linewidth]{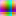}
&
\includegraphics[width=0.21\linewidth, height=0.21\linewidth]{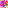}
&
\includegraphics[width=0.21\linewidth, height=0.21\linewidth]{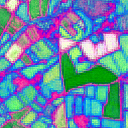}
\\
Time Series (T.S.) & Galileo (80m) & TerraMind (160m) & Tessera (10m) \\
\begin{subfigure}{0.223\linewidth}
\caption{Input}
\end{subfigure}
&
\multicolumn{2}{c}{
\begin{subfigure}{0.223\linewidth}
\caption{Multimodal Models}
\end{subfigure}}
&
\begin{subfigure}{0.223\linewidth}
\caption{Embedding Fields}
\end{subfigure}
\end{tabular}

\caption{{\bf Embedding Visualization.}
Embeddings of a PASTIS multimodal test tile (1.6 km\textsuperscript{2}) are projected using PCA, with the top three components mapped to RGB; colors are lightly harmonized across images for visualization.}
    \label{fig:featmap}
\end{figure}

\paragraph{Feature Maps.}
We visualize feature maps from different models in \cref{fig:featmap} using PCA projections.
Thanks to its controllable output resolution, \textsc{UniverSat} produces higher-resolution embeddings that preserve fine spatial structures (\eg, field boundaries and roads), compared to fixed-resolution models.
As observed by~\citet{ye2026any}, several models exhibit patterns consistent with \emph{positional collapse}~\citep{darcet2025cluster}, where dominant PCA components are largely driven by positional encodings.

\subsection{Ablation Study}
\label{sec:ablation}

We assess in \cref{tab:ablation} the impact of our main design choices in a simplified setting.
For all variants, we train a Tiny model (width $192$, $D=48$) on the full training corpus and evaluate it on one classification benchmark (m-Brick-Kiln) and two semantic segmentation benchmarks (Sen1Floods11 and PASTIS).

\begin{compactitem}
\item[\textsc{a}] {\bf UPE.}
We replace the UPE with $\vert \bM \vert$ modality-specific MLP projectors (\emph{UPE$\mapsto$MLPs}), as in a standard ViT.
This ties the model to sensors seen during training, prevents arbitrary patch sizes, and reduces the number of tokens processed by each encoder.
It leads to a marked performance drop, while increasing the parameter count by $58\%$ and removing the ability to handle unseen sensors. However, this variant is more efficient, with a $2\times$ reduction in training time.
Replacing the LFF atom embedding with a linear scalar embedding (\emph{LFF$\mapsto$Linear}) degrades performance, supporting a nonlinear embedding of atoms.
Reversing the aggregation order to collapse time before channels (\emph{Time before channels}) mostly affects PASTIS, which has long time series.
Finally, training only on the Sentinel-2 part of the corpus (\emph{S2 only training}) performs worse, even on the Sentinel-2-only Brick-Kiln: multimodal training causes no negative transfer and instead provides complementary supervision.

\item[\textsc{b}] {\bf UniverSat architecture.}
We remove the skip connection that enables cross-attention to sub-patch embeddings (\emph{No skip connection}).
This degrades performance on all three benchmarks, most strongly on Brick-Kiln ($-3.9$), showing that the skip connection helps preserve the fine-grained spatial features lost during patch-level aggregation.

We then disable resolution control by keeping the same patch size throughout the network (\emph{Fixed Output Res.}).
This degrades performance on all three benchmarks, most notably on PASTIS ($-3.7$) and Brick-Kiln ($-2.3$), showing that decoupling the output resolution from the input patch size benefits both segmentation and classification.
Finally, we replace ACA-based modality fusion with either late fusion (\emph{Late fusion}), where each modality is processed independently and the resulting embeddings are averaged, or mid-fusion (\emph{Average Pooling}), where $\ACA_M$ is replaced by average pooling over modalities right after the UPE.
Both cause a small performance drop, showing that ACA is an effective fusion mechanism. It adds little cost, as it operates on patch embeddings rather than atoms, whereas late fusion requires one pass per modality.

\item[\textsc{c}] {\bf Training objective.}
We remove the cross-modal contrastive loss applied to patch embeddings directly after the UPE ($\lambda_\text{con}=0$).
This substantially decreases performance, especially for multimodal segmentation.
We hypothesize that this auxiliary supervision stabilizes UPE training and encourages stronger alignment of sub-patch representations across modalities.
We also evaluate variants of our LM$_3$. Replacing the contrastive InfoNCE loss with a simple L2 loss in the frozen projection space results in a significant decrease in performance. Replacing LM$_3$ with a traditional MAE objective in pixel space also incurs a significant, but lesser, loss of performance. LM$_3$ is thus both easier to implement for massively multimodal training and more effective. Removing $\mathcal{L}_{\text{LM}^3}$ is not a meaningful ablation, as only the UPE would then be supervised.
\end{compactitem}
\begin{table}[]
    \centering
    \begin{minipage}{0.4\linewidth}
    \caption{{\bf Ablation Study.} We evaluate the impact of our key design choices in a simplified setting. }
    \label{tab:ablation}
    \end{minipage}
    \hspace{1mm}
    \begin{minipage}{0.5\linewidth}
    \resizebox{\linewidth}{!}{
    \begin{tabular}{ll c c c} 
\toprule
&& classification & \multicolumn{2}{c}{segmentation} \\
&& m-Brick-Kiln & Sen1flood11 & Pastis\\
  \midrule
 &Best Configuration  & \bf{92.4} & \bf{78.4}& \bf{36.4}\\\greyrule
 \textsc{a} & UPE$\mapsto$ MLPs & 91.9 & 77.6 & 21.5 \\
 \textsc{a} & LFF$\mapsto$ Linear atom emb. & 88.8 & 77.8 & 34.9 \\
 \textsc{a} & Time before channels & 91.2 & 76.8 & 25.2 \\
 \textsc{a} & S2 only training  & 87.0 &  -   & 29.3 \\\greyrule
 \textsc{b} & No skip connection & 88.5 & 76.9 & 34.0\\
 \textsc{b} & Fixed Output Res. & 90.1 & 77.5 & 32.7 \\
 \textsc{b} & Late modality fusion & - & - & 32.6\\
 \textsc{b} & Mid-fusion: Average Pooling & 92.1 & 76.8 & 32.7\\\greyrule
 \textsc{c} & No cross-modal contrast & 91.3 & 76.8 & 27.9 \\
 \textsc{c} & InfoNCE$\mapsto$ L2 & 69.8 & 69.7 & 10.7 \\
 \textsc{c} & Pixel MAE & 77.5 & 74.4 & 12.7 \\

\bottomrule
\end{tabular}

    }
    \end{minipage}
\end{table}
\paragraph{Limitations and Impact.}
\textsc{UniverSat} trades specialization for generality: in standard settings (\eg, VHR RGB or mono-temporal Sentinel-2), modality-specific models may be more accurate or efficient. Its computational overhead is moderate, and it is even faster on long time series, but memory use is typically higher (\cref{sec:app:computation_cost}): this is the cost of supporting heterogeneous inputs with one architecture. Our evidence for unseen sensors covers EnMAP and PlanetScope, whose characteristics lie within the pretraining gamut (roughly $0.2$--$250$\,m GSD, up to $140$ observations per year, and $1$--$500$ channels); we do not claim generalization to arbitrary sensors far outside it, such as UAV imagery or very dense temporal streams.
Axial aggregation assumes co-registered inputs: under strong misalignment across timestamps or modalities, the UPE may blur fine details, which standard co-registration can mitigate. It is also less seamless for non-optical sensors, which require learning a small modality encoding vector alongside the probe.
Beyond technical aspects, and as with any EO model, \textsc{UniverSat} may also enable large-scale monitoring capabilities, raising concerns around surveillance or misuse. %

\section{Conclusion}
\vspace{-0mm}
We introduced \textsc{UniverSat}, a transformer architecture for multimodal Earth Observation that replaces fixed patch projectors with a Universal Patch Encoder, enabling a single model to process multimodal inputs of any resolutions.
Through self-supervised training across 13 sensors and 7 heterogeneous datasets, \textsc{UniverSat} learns modality- and resolution-agnostic representations that transfer effectively to a wide range of downstream EO tasks.
Our results demonstrate that a single, shared backbone remains competitive with modality-specific foundation models, and surpasses them in several settings, while offering far greater input flexibility.

\begin{ack}
This work was supported by Hi! PARIS (ANR-23-IACL-0005), the ANR project SHARP (ANR-23-PEIA-0008) in the context of the PEPR IA, the French Space Agency PhD program (CNES), and the French National Institute of Geographic and Forest Information (IGN).
This work was granted access to the HPC resources of IDRIS under the allocations AD011014719R2, AD011014781R2, and A0200116178 made by GENCI.
We thank Jordi Inglada for inspiring discussions and valuable feedback.
\end{ack}

\FloatBarrier
\bibliographystyle{unsrtnat}
\bibliography{biblio}

\pagebreak
\setcounter{section}{0}
\renewcommand*{\theHsection}{App.\the\value{section}}
\renewcommand*{\thesection}{A.\arabic{section}}
\setcounter{figure}{0}
\renewcommand*{\theHfigure}{App.\thefigure}
\renewcommand\thefigure{A.\arabic{figure}}
\setcounter{table}{0}
\renewcommand*{\theHtable}{App.\thetable}
\renewcommand\thetable{A.\arabic{table}}
\maketitlesupplementary
\FloatBarrier

In this appendix, we provide details on compute cost and licensing (\cref{sec:app:details}), additional methodological insights into the Universal Patch Encoder (UPE) (\cref{sec:app:method}), and reproducibility details (\cref{sec:app:repro}).
We then report additional fine-tuning experiments (\cref{sec:app:experiments}), discuss computational cost (\cref{sec:app:computation_cost}) and information compression (\cref{sec:app:upe_compression}), and provide an exhaustive description of the training datasets with an extended comparison to competing methods (\cref{sec:app:data}).

\section{Compute Cost and License}
\label{sec:app:details}
\paragraph{License}
Code is released under the MIT License at \url{https://github.com/gastruc/UniverSat}, allowing unrestricted use, modification, and distribution. Pretrained weights are available on the Hugging Face Hub at \url{https://huggingface.co/g-astruc/UniverSat}. All datasets used in this work are publicly available; we adhere to their respective licenses and cite them appropriately in the paper and supplementary material.

\paragraph{Compute Resources}
All experiments were conducted on H100 GPUs. Training \textsc{UniverSat} in the self-supervised setting requires approximately 240 GPU-h. Linear probing experiments are lightweight, ranging from 1 to 10 GPU-h per dataset, for a total of 60 GPU-h.
The full project, including ablations and preliminary experiments, required approximately 30K GPU-h. Using standard estimates of carbon intensity computed with CodeCarbon\footnote{\url{https://codecarbon.io/}}, this corresponds to an approximate carbon footprint of 31 tons CO$_2$.

\section{Additional Methodological Details}
\label{sec:app:method}
We provide exact derivations of the Learnable Fourier Features, Metadata Encoding, and Axial Cross-Attention modules.

\subsection{Learnable Fourier Features (LFF)}
We use Fourier features followed by a learned projection, in the spirit of Learnable Fourier Features (LFF)~\citep{li2021learnable}, to lift scalar values
(\eg, radiance or wavelength) to a $D$-dimensional embedding
$
\mathrm{LFF} : \mathbb{R} \rightarrow \mathbb{R}^D.
$
For a scalar $u \in \mathbb{R}$,
\begin{align}
\mathrm{LFF}(u)
=
W\,\Big[\sqrt{2}\cos(\omega_k u + \phi_k)\Big]_{k=1}^{D/4},
\end{align}
where the frequencies $\omega_k \sim \mathcal{N}(0, (2\pi\beta)^2)$ and phases $\phi_k \sim \mathcal{U}(0, 2\pi)$ are sampled at initialization and kept frozen, and $W \in \mathbb{R}^{D \times D/4}$ is a learned linear projection. Following EDM2~\citep{karras2024analyzing}, the $\sqrt{2}$ factor yields unit-variance features. We use a bandwidth $\beta=2.5$ for atom values and $\beta=1$ for wavelengths.
When applied to a tensor, LFF  modules are applied element-wise.

\subsection{Metadata Encoding}
We consider a tensor $t \in \mathbb{R}^{X \times Y \times D}$, where $X$ is the axis to collapse, $D$ the feature dimension, and $Y$ aggregates all remaining dimensions. Patch $x$ is associated with its ground sampling distance (GSD) $\rho$, its acquisition times $\tau=\{\tau_t\}_{t=1}^T$, and  per-channel descriptors $b=\{b_c\}_{c=1}^C$: wavelength for optical data, categorical identifier otherwise. We add to $t$ some axis-specific metadata encodings, depending on the dimension $X$ being collapsed:
\begin{compactitem}
\item \textbf{Channel ($C$):} optical channels receive a dedicated LFF embedding of their wavelength (in \textmu m). Non-optical channels (\eg, radar polarizations or elevation products) cannot be described by a single scalar and are instead mapped to learned lookup embeddings.
\item \textbf{Time ($T$):} we apply 1D RoPE~\citep{heo2024rotary} to the day of the year of each acquisition. For multi-year series, the position is the day of the year plus $365$ times the number of years elapsed since the start of the sequence, which emphasizes seasonality while still distinguishing years.
\item \textbf{Space ($I$ and $S$):} we apply 2D RoPE relative positional encodings~\citep{heo2024rotary} at both spatial aggregation stages (pixels within sub-patches, then sub-patches within patches), with positions scaled by the GSD $\rho$ following Scale-MAE~\citep{reed2023scale}.
\end{compactitem}
Additive metadata embeddings are added to $t$ before computing keys and values, while RoPE is applied as a rotary transformation to the projected query and key vectors. This makes the ACA module both modality-aware and geometry-aware.

\subsection{Axial Cross-Attention (ACA)}
Directly projecting $e$ with an MLP is impractical as $C\!\times\!T\!\times\!I\!\times\!S\!\times\!D$ varies significantly across sensors. Using self-attention would be too expensive as the channel, pixel, and temporal dimensions can each reach the hundreds. Instead, we collapse one axis at a time with axial cross-attention \citep{ho2019axial}, see \cref{fig:aca}.

We consider an Axial Cross-Attention module $\ACA_X^\alpha$ where $X$ is the dimension to collapse, and $\alpha$ is the feature expansion factor. Let $t$ be a tensor of dimension $X\!\times\!Y\!\times \!D$ with $X$ the target dimension, $D$ the feature dimension, and $Y$ representing all remaining dimensions. For example, to collapse dimension $I$ of a tensor of dimension ${C\!\times\!T\!\times\!I\!\times S \times\!D}$, we would have $X=I$ and $Y=C\!\times \!T\!\times\!S$.

We first compute keys, queries, and values with RMSNorm ($\mathrm{N}$) as follows:
\begin{align}
K,V &= \psi_{\mathrm{k}}(\mathrm{N}(t)),\psi_{\mathrm{v}}(\mathrm{N}(t)) &\text{dim:}&\;\textstyle {X \times Y \times \alpha D}\\
Q &= \psi_{\mathrm{q}}\big(\big[\mathrm{mean}_X(\mathrm{N}(t)),\,\mathrm{max}_X(\mathrm{N}(t))\big]\big) &\text{dim:}&\;\textstyle  {Y \times \alpha D}~,
\end{align}
where $\psi_{\mathrm{k}}$, $\psi_{\mathrm{v}}$ are linear layers mapping $D$ to the expanded feature axis $\alpha D$, $\mathrm{mean}_X$ and $\mathrm{max}_X$ pool along the $X$ axis, $[\cdot,\cdot]$ denotes concatenation, and $\psi_{\mathrm{q}}: 2D \mapsto \alpha D$. Queries and keys are further normalized per head (QK-norm).
We perform gated cross-attention along axis $X$ only, then apply a residual feed-forward network
$\mathrm{FF}$ with RMSNorm:
\begin{equation}
\label{eq:aca}
\mathrm{ACA}_{X}^\alpha(t)=\mathrm{FF}\left(\pool^\alpha_X (t)+\gamma_{CA}\,\mathrm{softmax}_X \left(\frac{Q \otimes_D K}{\sqrt{\alpha D}}\right) \otimes_X V\right)~,
\end{equation}
where $\mathrm{softmax}_X$ is on axis $X$, $\pool^\alpha_X$ is a max pooling along $X$ for $\alpha=1$ and the concatenation of mean and max pooling for $\alpha=2$, and $\gamma_{CA}$ is a LayerScale parameter initialized at $10^{-6}$ (the residual branch of $\mathrm{FF}$ uses the same scheme). We denote by $\otimes_D$ the vector product in axis $D$ with broadcasting on dimension $X\!\times\!Y$, and $\otimes_X$ the product in axis $X$ with broadcasting on dimension $Y\!\times\!D$:
\begin{align}
[Q \otimes_D K]_{x,y}
=
\sum_d Q_{y,d} K_{x,y,d} &\quad&
[A \otimes_X V]_{y,d}
=
\sum_x A_{x,y} V_{x,y,d}~.
\end{align}
Since each query attends only along axis $X$, the cost is \emph{linear} in the number of atomic tokens in each patch.

\subsection{Scale Augmentation and Masking}
We consider a multimodal tile $x=\{x_p^m\}_{p \in \bP,\, m \in \bM}$ composed of patches $\bP$ observed in modalities $\bM$.
We randomly sample both the input patch size and the target output resolution from dataset-specific presets, encouraging robustness to spatial scale.
We then apply a compositional masking operator:
\begin{compactitem}
\item \textbf{Modality dropping.} Randomly drop modalities for the entire tile with probability $30\%$, ensuring that at least one remains. The retained modalities are denoted $\bM^\star$.
\item \textbf{Time dropping.} For time-series modalities, randomly drop $50\%$ of timestamps and, with probability $10\%$, retain a single timestamp to emulate mono-date observations.
\item \textbf{Channel dropping.} For each retained modality with $N$ channels, drop a fraction
$
\psi(N)={1}/{(1+\sqrt{5/N})},
$
so that spectrally dense sensors are thinned more aggressively. Dropping is applied in two contiguous channel blocks.
\item \textbf{Patch dropping.} Drop a random subset $\bD \subset \bP$ of $75\%$ of the patches, and denote the visible patches by $\bP^\star=\bP\setminus \bD$.
\end{compactitem}
Overall, approximately $90\%$ of input atoms are removed.

\subsection{Linear Multimodal Masked Modeling (LM$_3$)}

We adapt the LMIM-Lite objective~\citep{wei2024towards,herzog2025olmoearth} to a multimodal and multitemporal setting.
The goal is to predict representations of masked patches in a fixed latent space defined by frozen random projections.

\paragraph{Random Projection Targets}
For each modality $m$, we define a frozen linear projection applied to single-time-step slices:
\begin{align}
\phi^{\text{rand}}_{m}: \mathbb{R}^{C \times H \times W} \rightarrow \mathbb{R}^{D}.
\end{align}
For optical modalities, varying patch sizes are handled via kernel interpolation, following FlexiViT~\citep{beyer2023flexivit}.
For SAR data, where spatial resampling is less straightforward, we use separate projections for each patch size.
The projections $\phi^{\text{rand}}_{m}$ are randomly initialized at the beginning of training and kept frozen thereafter.

\paragraph{Temporal Sampling}
A key challenge is to balance spatial and temporal training signals.
Using all timestamps may lead the model to exploit trivial temporal cues (\eg, acquisition dates) instead of spatial content, while ignoring temporal variation degrades representation quality.
To address this, we select a small set of $K=4$ timestamps per tile, sampled at random among those with a cloud density below $30\%$ when cloud masks are available.
Each masked patch $p \in \bD$ is assigned a target timestamp $\tau(p)$ from this set.

\paragraph{Prediction}
We define a predictor network $\phi^{\text{pred}}$, composed of a stack of self-attention blocks.
It takes as input the visible patch embeddings $\{z_q\}_{q \in \bP^\star}$ together with learned mask tokens for masked patches $p \in \bD$.
We apply spatial RoPE positional encodings to all tokens, and inject absolute temporal encodings into masked tokens to represent their assigned timestamps $\tau(p)$.
The predictor produces representations for masked patches at their target times:
\begin{align}
\{ z_p^{\tau(p)} \}_{p \in \bD}
=
\phi^{\text{pred}}\big(\{z_q\}_{q \in \bP^\star}\big).
\end{align}

Modality-specific linear heads $\{\phi^{\text{head}}_m\}_{m \in \bM}$ map these embeddings to the target space:
\begin{align}
h_p^{m,\tau(p)} = \phi^{\text{head}}_m\big(z_p^{\tau(p)}\big).
\end{align}

\paragraph{Loss}
We define a contrastive loss between predicted embeddings and random projection targets of the masked inputs at the corresponding timestamps, using temperature $\tau_{\mathrm{LM}^3}$:
\begin{align}
\mathcal{L}_{\text{LM}^3}
=
-\sum_{m \in \bM}
\sum_{p \in \bD}
\log
\frac{
\exp\left(
\frac{1}{\tau_{\mathrm{LM}^3}}
\left\langle
h_p^{m,\tau(p)},
\phi^{\text{rand}}_{m}\!\left(x_p^m[\tau(p)]\right)
\right\rangle
\right)
}{
\sum_{q \in \bD}
\exp \left(
\frac{1}{\tau_{\mathrm{LM}^3}}
\left\langle
h_p^{m,\tau(p)},
\phi^{\text{rand}}_{m}\!\left(x_q^m[\tau(q)]\right)
\right\rangle
\right)
}.
\end{align}

This objective combines masked prediction in latent space with a contrastive training signal.
It avoids collapse without requiring momentum teachers, while coupling information across modalities and time.

\section{Reproducibility Details}
\label{sec:app:repro}
\providecommand{\missingdetail}[1]{\textcolor{red}{\emph{#1}}}

We summarize the implementation and evaluation details required to reproduce the experiments.

\subsection{Model Configurations}
\begin{table}[!htbp]
\centering
\small
\setlength{\tabcolsep}{4pt}
\caption{Model configuration details required for reproducing \textsc{UniverSat}-Tiny and \textsc{UniverSat}-B.}
\label{tab:app:repro_model}
\begin{tabularx}{\linewidth}{@{}p{0.25\linewidth}cc@{}}
\toprule
\textbf{Item} & \textbf{\textsc{UniverSat}-Tiny} & \textbf{\textsc{UniverSat}-B} \\
\midrule
Usage & Ablation study in \cref{tab:ablation} & Main benchmark tables \\
Atomic embedding size $D$ & 48 & 96 \\
UPE patch embedding & 192 & 768 \\
UPE expansion $\alpha$ ($I$, $C$, $T$, $S$) & 1, 2, 2, 2 & 2, 2, 2, 2 \\
UPE sequence & \multicolumn{2}{c}{Pixel, Channel, Time, Sub-patch} \\
Spatial Transformer blocks & SA$\times$6 & SA$\times$12 \\
Attention heads: UPE + encoder & 8&12 \\
Attention heads: predictor & 8&12 \\
FFN expansion ratio & \multicolumn{2}{c}{4} \\
Normalization & \multicolumn{2}{c}{Pre-norm RMSNorm, QK-norm} \\
Gating (MLP and Attention) & \multicolumn{2}{c}{True} \\
QKV bias (ACA, CA and SA) & \multicolumn{2}{c}{False} \\
Dropout & \multicolumn{2}{c}{dropout 0.0; attention dropout 0.0} \\
Stochastic depth & \multicolumn{2}{c}{stochastic depth 0.0} \\
Register tokens & \multicolumn{2}{c}{4} \\
UPE register tokens & \multicolumn{2}{c}{2 (time axis), 4 (sub-patch axis)} \\
Predictor depth & \multicolumn{2}{c}{depth 8} \\
Predictor time subset size  & \multicolumn{2}{c}{4} \\
Predictor register & \multicolumn{2}{c}{4} \\
Predictor dimensions & 192 & 768 \\
Random targets & \multicolumn{2}{c}{Frozen linear projection; FlexiViT kernel at $32\times32$ or lower (SAR: one per patch size)} \\
Prediction heads $\phi^{\text{head}}_m$ & \multicolumn{2}{c}{Linear, one per modality} \\
\bottomrule
\end{tabularx}

\end{table}
\FloatBarrier

\subsection{Self-Supervised Pretraining}
\begin{table}[!htbp]
\centering
\small
\setlength{\tabcolsep}{4pt}
\caption{Self-supervised pretraining hyperparameters and optimization details.}
\label{tab:app:repro_pretraining}
\begin{tabularx}{\linewidth}{@{}p{0.28\linewidth}X@{}}
\toprule
\textbf{Item} & \textbf{Value / description} \\
\midrule
Supervision & Self-supervised only; no labels are used during pretraining. \\
Time & Full pretraining requires approximately 240 GPU-h. \\
GPUs        & 2 nodes $\times$ 4 GPUs per node = 8 GPUs \\
GPU memory & H100 80GB \\
Precision & bf16 mixed precision \\
Optimizer type & AdamW \\
Optimizer LR & $5\times10^{-4}$ \\
Optimizer weight decay & $10^{-3}$ \\
Optimizer epsilon and betas & $\epsilon=10^{-5}$, betas use PyTorch AdamW defaults \\
LR schedule & 10 warmup epochs, then cosine annealing to $10^{-6}$ \\
Gradient accumulation & 2\\
Training length& 100k steps \\
Final loss & $\mathcal{L}_{\mathrm{LM}^3}+\lambda_{\mathrm{con}}\mathcal{L}_{\mathrm{con}}$. \\
Loss weights & $\lambda_{\mathrm{con}}=1.0$ \\
LM$^3$ temperature $\tau_{\mathrm{LM}^3}$ & 0.1 \\
Cross-modal contrast temperature & 0.2 \\
\bottomrule
\end{tabularx}

\end{table}
\FloatBarrier

\begin{table}[!htbp]
\centering
\scriptsize
\setlength{\tabcolsep}{3pt}
\caption{Modality-level preprocessing configuration: native resolution, number of input bands or descriptors, sub-patch factor, and datasets using each modality.}
\label{tab:app:repro_modalities}
\begin{tabularx}{\linewidth}{@{}p{0.20\linewidth}p{0.14\linewidth}p{0.12\linewidth}p{0.14\linewidth}X@{}}
\toprule
\textbf{Modality} & \textbf{Resolution (m)} & \textbf{Bands} & \textbf{Sub-patch size (px)} & \textbf{Datasets} \\
\midrule
spotRGBN & 1.6 & 4 & 2 & FLAIR-Hub. \\
aerialflair & 0.2 & 4 & 8 & FLAIR-Hub. \\
s2flair & 12.8 & 10 & 1 & FLAIR-Hub. \\
s1flair & 12.8 & 3 & 1 & FLAIR-Hub. \\
dem & 0.2 & 2 & 10 & FLAIR-Hub. \\
spot & 1 & 3 & 10 & PASTIS-HD. \\
s2 & 10 & 10 & 1 & PASTIS-HD, Planted, TSAI-TS, S2NAIP-Urban. \\
s1 & 10 & 3 & 1 & PASTIS-HD, Planted, TSAI-TS, S2NAIP-Urban. \\
l7 & 30 & 6 & 1 & Planted. \\
alos & 30 & 3 & 1 & Planted. \\
modis & 250 & 7 & 1 & Planted. \\
aerial & 0.2 & 4 & 10 & TSAI-TS. \\
naip & 1.25 & 4 & 10 & S2NAIP-Urban. \\
l8 & 10 & 11 & 1 & S2NAIP-Urban. \\
EO1 & 30 & 175 & 1 & HyperGlobal. \\
GF5 & 30 & 150 & 1 & HyperGlobal. \\
enmap & 30 & 202 & 1 & SpectralEarth (evaluation only). \\
rgbneon & 0.1 & 3 & 20 & EarthView. \\
ndemneon & 1 & 1 & 2 & EarthView. \\
neon & 1 & 369 & 2 & EarthView. \\
\bottomrule
\end{tabularx}

\end{table}
\FloatBarrier

\begin{table}[!htbp]
\centering
\small
\setlength{\tabcolsep}{3pt}
\caption{Dataset-level spatial configuration: image size from the config, input and target output scales, and modalities used for each dataset.}
\label{tab:app:repro_dataset_scales}
\begin{tabularx}{\linewidth}{@{}p{0.14\linewidth}p{0.14\linewidth}p{0.16\linewidth}p{0.16\linewidth}p{0.12\linewidth}X@{}}
\toprule
\textbf{Dataset} & \textbf{Image size ($\times$10\,m)} & \textbf{Input scales ($\times$10\,m)} & \textbf{Output scales ($\times$10\,m)} & \textbf{Dataset weight} & \textbf{Modalities} \\
\midrule
FLAIR-Hub & 10.24 & 1.28, 2.56 & 0.64, 1.28 & 2 & spotRGBN, aerialflair, s2flair, s1flair, dem. \\
PASTIS-HD & 128 & 4, 8, 16 & 2, 4 & 2 & spot, s2, s1. \\
Planted & 12 & 3 & 3 & 1 & s2, s1, l7, alos, modis. \\
TSAI-TS & 6 & 2, 1 & 1 & 1 & aerial, s2, s1. \\
S2NAIP-Urban & 64 & 16, 4, 8 & 2, 4 & 4 & naip, l8, s2, s1. \\
HyperGlobal & 192 & 12, 6 & 6, 3 & 2 & EO1, GF5. \\
EarthView & 6.4 & 0.8, 1.6 & 0.4, 0.8 & 4 & rgbneon, ndemneon, neon. \\
\bottomrule
\end{tabularx}

\end{table}
\FloatBarrier

\section{Additional Experiments}
\label{sec:app:experiments}

\subsection{End-to-End Fine-Tuning}
\label{sec:app:finetuning}
Our main protocol uses frozen-backbone probing, following Galileo~\citep{tseng2025galileo} and OlmoEarth~\citep{herzog2025olmoearth}.
To verify that the UPE does not cause optimization instability, we also fine-tune \textsc{UniverSat}-B end-to-end on Sen1Floods11 and BurnScars.
The model trains stably and reaches strong performance, setting a new state of the art on Sen1Floods11 and approaching the best models on BurnScars (\cref{tab:app:finetuning}).

\begin{table}[h]
\centering
\caption{\textbf{End-to-end fine-tuning.} mIoU of fully fine-tuned models.}
\label{tab:app:finetuning}
\begin{tabular}{lcc}
\toprule
Model & Sen1Floods11 & BurnScars \\
\midrule
AnySat~\citep{astruc2024anysat} & 77.4 & \bf 90.6 \\
RemoteCLIP & -- & 65.1 \\
Prithvi~\citep{jakubik2310foundation} & -- & 86.9 \\
Prithvi v2~\citep{szwarcman2024prithvi} & -- & 90.5 \\
Panopticon~\citep{waldmann2025panopticon} & 79.1 & -- \\
CROMA~\citep{fuller2023croma} & 79.4 & -- \\
TerraMind~\citep{jakubik2025terramind} & 79.5 & -- \\
OlmoEarth~\citep{herzog2025olmoearth} & 79.8 & -- \\
\greyrule
\textsc{UniverSat}-B, linear probing & 80.1 & 81.5 \\
\textsc{UniverSat}-B, fine-tuning & \bf 80.2 & 89.6 \\
\bottomrule
\end{tabular}
\end{table}

\section{Limitations}
\subsection{Computational Efficiency}
\label{sec:app:computation_cost}

To quantify UniverSat’s overhead, we compare it to a minimal ViT-style baseline in which the UPE is replaced by independent linear projectors for each modality, patch size, and resolution, resulting in a token for each patch, modality, and timestep. This baseline is particularly computationally efficient but also highly impractical: it requires a separate projector for every input modality and patch size (42 projectors), cannot be applied to unseen configurations, and performs substantially worse, \eg, by approximately 15 points on PASTIS (\emph{UPE$\mapsto$MLPs} in \cref{tab:ablation}). This suggests that the cheaper interface fails to learn shared geospatial representations across heterogeneous inputs.

We now report throughput and memory for representative input types on A100, using images of 128×128 for S1/S2 time series with patch size 4 and images of 512×512 with patch size 64 for aerial photography. 

\begin{table*}[t]
    \centering
    \begin{tabular}{l c l c c c c c c}
\toprule
    \multirow{2}{*}{Data} & \multirow{2}{*}{Timesteps} & \multirow{2}{*}{Model} & \multicolumn{2}{c}{Inference} & \multicolumn{2}{c}{Training} & \multicolumn{2}{c}{Memory} \\
     &  &  & \multicolumn{2}{c}{(samples/s)} & \multicolumn{2}{c}{(samples/s)} & \multicolumn{2}{c}{(GB VRAM)} \\
    \cmidrule(lr){4-5} \cmidrule(lr){6-7} \cmidrule(lr){8-9}
    \multirow{2}{*}{S2} & \multirow{2}{*}{40}  & MultiMLP baseline & 2.1  & - & 6.9 & - & 4.6  & - \\
                        &                      & UniverSat         & 2.0  & $\times$0.95 & 2.7 & $\times$0.39 & 12.2 & $\times$2.7 \\\greyrule
    \multirow{2}{*}{S1} & \multirow{2}{*}{140} & MultiMLP baseline & 0.21 & - & 1.8 & - & 11.8 & - \\
                        &                      & UniverSat         & 1.7  & $\times$8.1  & 1.8 & $\times$1.0  & 12.7 & $\times$1.1 \\\greyrule
    \multirow{2}{*}{Aerial 20cm} & \multirow{2}{*}{-} & MultiMLP baseline & 130  & - & 26  & - & 1.8  & - \\
                        &                      & UniverSat         & 79   & $\times$0.61 & 14  & $\times$0.54 & 3.7  & $\times$2.1 \\
\bottomrule
\end{tabular}

    \caption{\textbf{Computational cost.} Throughput and memory of \textsc{UniverSat} and of a ViT-style baseline with one linear projector per modality and patch size (MultiMLP), measured on an A100 GPU. Ratios are relative to the baseline.}
    \label{tab:app:complexity}
\end{table*}

UniverSat is slower and more memory-intensive for short time series and high-resolution aerial imagery, but its overhead remains moderate relative to the added flexibility. Memory usage is generally higher, but can be controlled through the input patch size and target output resolution.

These results clarify the intended trade-off. UniverSat is not designed to be cheaper than a specialized encoder optimized for one fixed input format. Instead, it provides a single native-resolution interface that avoids resampling, supports unseen configurations, and yields stronger representations across heterogeneous EO inputs at a measurable computational cost.

\subsection{UPE Compression of Information}
\label{sec:app:upe_compression}

Due to its flexibility, the UPE maps heterogeneous patches to fixed-width representations which can lead to discarding some raw information. It is a representation encoder, not an invertible reconstruction model. However, it does not collapse the complete patch into a single vector: it outputs both a global patch embedding and a set of sub-patch embeddings used by the high-resolution skip connection.

The effective compression rate depends on the physical patch extent and on the spatial, spectral, and temporal density of the sensor. Across our input configurations, it ranges from an expansion of 0.02× for sparse ALOS inputs to a compression of 369× for the dense NEON hyperspectral time series, with a geometric average of 1.56× when removing NEON. Importantly, the patch extent is selected at runtime rather than fixed by the architecture, allowing the user to control this trade-off.

Such compression is not unique to UPE: a standard ViT also maps each input patch to a fixed-width token. The main difference is that UPE supports heterogeneous input shapes with shared weights and preserves sub-patch embeddings, which mitigate spatial information loss and retain fine-scale features for dense prediction.

The largest compression rates occur for very-high-resolution and hyperspectral inputs, where adjacent pixels or spectral bands are often strongly redundant. Our strong frozen-feature performance on hyperspectral benchmarks suggests that the UPE preserves substantial task-relevant information even in this regime, although we do not claim that it retains every raw measurement.

\clearpage
\begingroup
\raggedbottom
\section{Data and Models}
\label{sec:app:data}

We provide here an extended comparison with recent models (\cref{tab:app:models}) and the datasets used to train \textsc{UniverSat} (\cref{tab:app:datasets}).

\begin{table}[H]
    \centering
    \definecolor{colVHR}{RGB}{46,178,164}%
\definecolor{colOpt}{RGB}{254,170,94}%
\definecolor{colRadar}{RGB}{110,150,230}%
\definecolor{colHyper}{RGB}{214,106,205}%

\begin{tabular}{l x{5mm}y{5mm}x{5mm}y{5mm}x{5mm}y{5mm} x{10mm}y{10mm}x{10mm} x{20mm}}
\toprule
 & \multicolumn{6}{c}{training modalities}
 & \multicolumn{3}{c}{supports unseen resolution}
 & \multirow{2}{*}[-2mm]{\cellcolor{gray!10}\makecell{feature\\resolution}}
 \\ \cmidrule(lr){2-7}\cmidrule(lr){8-10}
 & \cellcolor{colVHR!60}\faCamera & \cellcolor{colOpt!60} \faVideo & \cellcolor{colRadar!60}\faRadar{1} & \cellcolor{colDEM!60}\faElevation{1} & \cellcolor{colHyper!60}\faHyper{1}& \faPenNib
 & spat. & temp. & spec. & 
 \\\midrule

DINOv3~\citep{simeoni2025dinov3} & \applycolorA{1} & & & & & & & & & \faPatch{1}\\
\rowcolor{gray!10}SatMAE~\citep{cong2022satmae} & & \applycolorB{1}  &  &  &   & & & & & \faPatch{1}\\
CROMA~\citep{fuller2023croma} &  \applycolorA{1} &  &  \applycolorC{1} & & & & & & & \faPatch{1} \\
\rowcolor{gray!10} DOFA~\citep{xiong2024dofa} & \applycolorA{3} &  & \applycolorC{1}  & &\applycolorF{1} & & \faCheck & & \faCheck  &\faPatch{1}\\
Atomizer~\citep{de2025atomizer} & \applycolorA{1} & & & &  &  & & & \faCheck &\faPatch{1} \\
\rowcolor{gray!10}Prithvi v2~\citep{szwarcman2024prithvi} & & \applycolorB{1} & & & & & & & &\faPatch{1} \\
DUNIA~\citep{fayad2025dunia} & & \applycolorB{1}   & & \applycolorE{1}  & &  & & &\faCheck & \faPatch{1}\\
\rowcolor{gray!10}MAESTRO~\citep{labatie2025maestro} &\applycolorA{1}  & \applycolorB{1}  &  \applycolorD{1}  & & & & & \faCheck & &\faPatch{1}\\
SkySense v2~\citep{zhang2025skysense} &\applycolorA{1} &  \applycolorB{1} & \applycolorD{1}  & & & & & \faCheck & &\faPatch{1}\\
\rowcolor{gray!10}PRESTO~\citep{tseng2023lightweight} &  & \applycolorB{1}  &  \applycolorD{1} & \applycolorE{1} & & \bf \applycolorG{6} &  & \faCheck & &\faPixel{1}\\
EarthView~\citep{velazquez2025earthview} & \applycolorA{3}  &   & \applycolorC{1}  &\applycolorF{1} & & & & \faCheck & &\faPatch{1}\\
\rowcolor{gray!10}Galileo~\citep{tseng2025galileo} &  & \applycolorB{1}  &\applycolorD{1} & \applycolorE{1}& & \bf \applycolorG{6} &  \faCheck  & \faCheck & &\faPatch{1}\\
Smarties~\citep{sumbul2025smarties} & \applycolorA{1} & & \applycolorC{1}& & & & & & \faCheck & \faPatch{1}\\
\rowcolor{gray!10}TerraMind~\citep{jakubik2025terramind} & \applycolorA{1} & & \applycolorC{1} &\applycolorE{1} & \applycolorG{1} & & \faCheck & & & \faPatch{1}\\ %
Scale-MAE~\citep{reed2023scale} & \applycolorA{1} & & & & & &  \faCheck & & & \faPatch{1} \\
\rowcolor{gray!10}TerraFM~\citep{danish2025terrafm} & \applycolorA{1} & & \applycolorC{1} & & \applycolorG{2} & & & & & \faPatch{1} \\
Panopticon~\citep{waldmann2025panopticon} & \bf \applycolorA{5} & & \applycolorC{1} & & \applycolorF{1} & & & & \faCheck & \faPatch{1} \\
\rowcolor{gray!10}AlphaEarth~\citep{brown2025alphaearth} & &\applycolorB{3} & \applycolorD{1} & \applycolorE{2} & & \applycolorG{5} &  & \faCheck & & \faPixel{1}\\
OmniSat~\citep{astruc2024omnisat} & \applycolorA{2} & \applycolorB{2} & \applycolorD{1} & \applycolorE{1} & & & & \faCheck & &\faFullImage{1} \\

\rowcolor{gray!10}FoMo~\citep{bountos2025fomo}& \applycolorA{4} & \applycolorB{2} &\applycolorD{1} & \applycolorE{1} &  & & & \faCheck & \faCheck & \faPatch{1} \\ 
Ramen~\citep{houdre2025ramen} & \applycolorA{1} & \applycolorB{1} & \applycolorC{1} & \applycolorD{1} & & & \faCheck & \faCheck & \faCheck & \faAnyRes{1} \\ 
\rowcolor{gray!10}AnySat~\citep{astruc2024anysat} &  \applycolorA{2} &  \applycolorB{4}  &\bf \applycolorD{2} &  \applycolorE{1} & & & \faCheck & \faCheck & & \faPixel{1}\\
\greyrule 
UniverSat &  \applycolorA{4}  & \bf \applycolorB{4} & \bf \applycolorD{2} & \bf \applycolorE{2} & \applycolorG{3}  &   & \faCheck& \faCheck  & \faCheck & \faAnyRes{1}
\\ \bottomrule
\end{tabular}
\begin{tabular}{c}
\begin{tabular}{r@{ : }l r@{ : }l r@{ : }l r@{ : }l r@{ : }l r@{ : }l}
\faCamera/\faVideo & optical snapshot/ time series & \faRadar{1} & radar & \faElevation{1} & elevation & \faHyper{1} & hyperspectral & \faPenNib & labels 
\end{tabular}\\
\begin{tabular}{r@{ : }l r@{ : }l r@{ : }l r@{ : }l  }
\faFullImage{1} & image & \faPatch{1} & patch &  \faPixel{1} & pixel &  \faAnyRes{1} & any resolutions
\end{tabular}
\end{tabular}

\caption{\textbf{Multimodal EO Foundation Models.}
For each model, we list the training modalities, whether unseen spatial/temporal/spectral configurations are handled \emph{without input resampling}, and the feature-map granularity.
\textsc{UniverSat} supports the broadest modality mix, handles unseen configurations, and offers flexible output resolution---all with a single set of weights.}
    \label{tab:app:models}
\end{table}

\begin{table}[H]
    \centering
    \small{
    \begin{tabular}{lcccccccc}
    \toprule
    \multirow{3}{*}[-0.8mm]{Dataset} & \multirow{3}{*}[-0.8mm]{Labels} & \multirow{3}{*}[-0.8mm]{Extent} & \multirow{3}{*}[-0.8mm]{Modalities} & \multicolumn{3}{c}{Resolution} &\multirow{2}{*}[-0.8mm]{\# atoms}&\\\cmidrule(l{8pt}r{8pt}){5-7} 
    & &&&Spat. & Temp. & Spect. 
\\
&&&&m&img/yr&chan.&G\aa\\\midrule
        \multirow{3}{*}{TSAI-TS %
        } &
        \multirow{3}{*}{\makecell{Tree species  \\  classification  \\ multilabel}
        }
        & 
        \multirow{3}{*}{\makecell{50K\\60$\times$60 m}}
         & Aerial VHR  & 0.20& 1 & 4 & 18 \\ %
        &&& S1 & 10& 10-70 & 3 & 0.2\\
        &&& S2 & 10& 10-70 & 10 & 0.7\\\greyrule
        \multirow{3}{*}{PASTIS-HD %
        } &
        \multirow{3}{*}{\makecell{Crop type  \\  segmentation  \\ dense}
        }
        &
        \multirow{3}{*}{\makecell{ 2433 \\1280$\times$1280 m}} & SPOT6/7  & 1m$^\dagger$
        & 1 & 4 & 16\\
        &&& S1 & 10& 140 & 3 & 17 \\
        &&& S2 & 10& 38-61 & 10 & 18 \\\greyrule
       
        \multirow{5}{*}{FLAIR-Hub%
        } &
        \multirow{5}{*}{\makecell{Land cover  \\  segmentation}
        }
        &
         \multirow{5}{*}{\makecell{241K \\ 102.4$\times$102.4 m}} &
         Aerial VHR  & 0.2& 1 & 4 & 253 \\
         &&& DSM+nDSM& 0.2& 1 & 1 & 63\\
         &&& SPOT6 & 1.5& 1 & 4 & 3.9\\
        &&& S1 & 10& 0-122 & 2 & 9.1\\
        &&& S2 & 10& 20-146 & 10 & 21
        \\\greyrule
        \multirow{5}{*}{Planted %
        } &
        \multirow{5}{*}{\makecell{Tree species  \\  classification}
        }
        &
        \multirow{5}{*}{\makecell{1.3M\\  120$\times$120 m}} & S2 & 10& 8 %
         & 10 & 15\\
        &&& S1 & 10& 8 %
        & 3 & 4.5\\
        &&& Landsat 7 & 30& 20 %
        & 3 & 1.2\\
        &&& ALOS-2 & 30& 4 %
        & 3 & 0.3\\
        &&& MODIS & 250& 60 %
        & 7 & 0.6\\\greyrule
        
        &
        \multirow{4}{*}{\makecell{Land cover  \\  segmentation  \\ dense}
        }
         &
         \multirow{4}{*}{\makecell{ 515K  \\ 640$\times$640 m}} 
        
         & NAIP & 1.25& 1 & 4 & 540 \\
       \multirow{1}{*}{S2NAIP-} &&& S2 & 10& 16-32  & 10 & 506 \\
        \multirow{1}{*}{Urban %
        }
        &&& S1 & 10& 2-8 & 3 & 38\\
        &&& Landsat 8/9 & 10m$^\dagger$ & 4 & 8 & 68\\\greyrule
       \multirow{2}{*}{HyperGlobal} && 450K & Gaofen-5 & 30& 1 & 150 & 151 \\
       && 1920$\times$1920 m & EO 1 & 30& 1 & 175&
       143\\\greyrule
        \multirow{3}{*}{EarthView}&& \multirow{3}{*}{\makecell{ 35K  \\ 64 $\times$64  m }} & UHR aerial RGB & 0.1& 1 & 3 & 44\\
       &&& NIS & 1& 3 & 396 & 173 \\
       &&& LiDAR & 1& 3 & 1 & 0.4\\
      \bottomrule
    \end{tabular}
}

    \caption{\textbf{Pretraining Datasets Overview.} Summary of the seven heterogeneous datasets used to train \textsc{UniverSat}. The table details the semantic labels, spatial extent, and sensor modalities for each dataset, along with their respective spatial (m), temporal (images/year), and spectral (channels) resolutions, and the total volume of atomic tokens (G\aa{}).}
    \label{tab:app:datasets}
\end{table}
\clearpage
\endgroup

\end{document}